\documentclass[10pt,twocolumn,letterpaper]{article}

\usepackage{iccv}              

\definecolor{iccvblue}{rgb}{0.21,0.49,0.74}
\usepackage[pagebackref,breaklinks,colorlinks,allcolors=iccvblue]{hyperref}

\def\paperID{8440} 
\def\confName{ICCV}
\def\confYear{2025}

\title{MobileIE: An Extremely Lightweight and Effective ConvNet for Real-Time Image Enhancement on Mobile Devices}

\author{
Hailong Yan$^{1}$,
Ao Li$^{1}$,
Xiangtao Zhang$^{1}$,
Zhe Liu$^{1}$,
Zenglin Shi$^{2}$,
Ce Zhu$^{1}$,
Le Zhang$^{1}$\thanks{Corresponding author.} \\
$^{1}$UESTC \quad
$^{2}$Hefei University of Technology \\
{\tt\small yanhailong@std.uestc.edu.cn, lezhang@uestc.edu.cn}
}

\begin{document}
\maketitle
\begin{abstract}
Recent advancements in deep neural networks have driven significant progress in image enhancement (IE). However, deploying deep learning models on resource-constrained platforms, such as mobile devices, remains challenging due to high computation and memory demands. To address these challenges and facilitate real-time IE on mobile, we introduce an extremely lightweight Convolutional Neural Network (CNN) framework with around 4K parameters. Our approach integrates re-parameterization with an Incremental Weight Optimization strategy to ensure efficiency. Additionally, we enhance performance with a Feature Self-Transform module and a Hierarchical Dual-Path Attention mechanism, optimized with a Local Variance-Weighted loss. With this efficient framework, we are the first to achieve real-time IE inference at up to 1,100 frames per second (FPS) while delivering competitive image quality, achieving the best trade-off between speed and performance across multiple IE tasks. The code will be available at https://github.com/AVC2-UESTC/MobileIE.git.
\end{abstract}

\section{Introduction}
\label{sec:intro}

Image enhancement (IE) aims to restore images degraded by factors such as camera limitations \cite{zamir2020learning}, poor lighting \cite{liu2024ntire}, or challenging environments (e.g., underwater \cite{peng2023u}). With the growing prevalence of smart devices and embedded systems, real-time image enhancement has become essential in applications like mobile devices. These scenarios require not only high-quality output but also real-time processing, demanding efficient image enhancement within the constraints of limited computational resources \cite{chen2022simple,cui2023focal,cui2023image,cui2024revitalizing}.

\begin{figure}[h]
\centering
\includegraphics[width=0.9\linewidth]{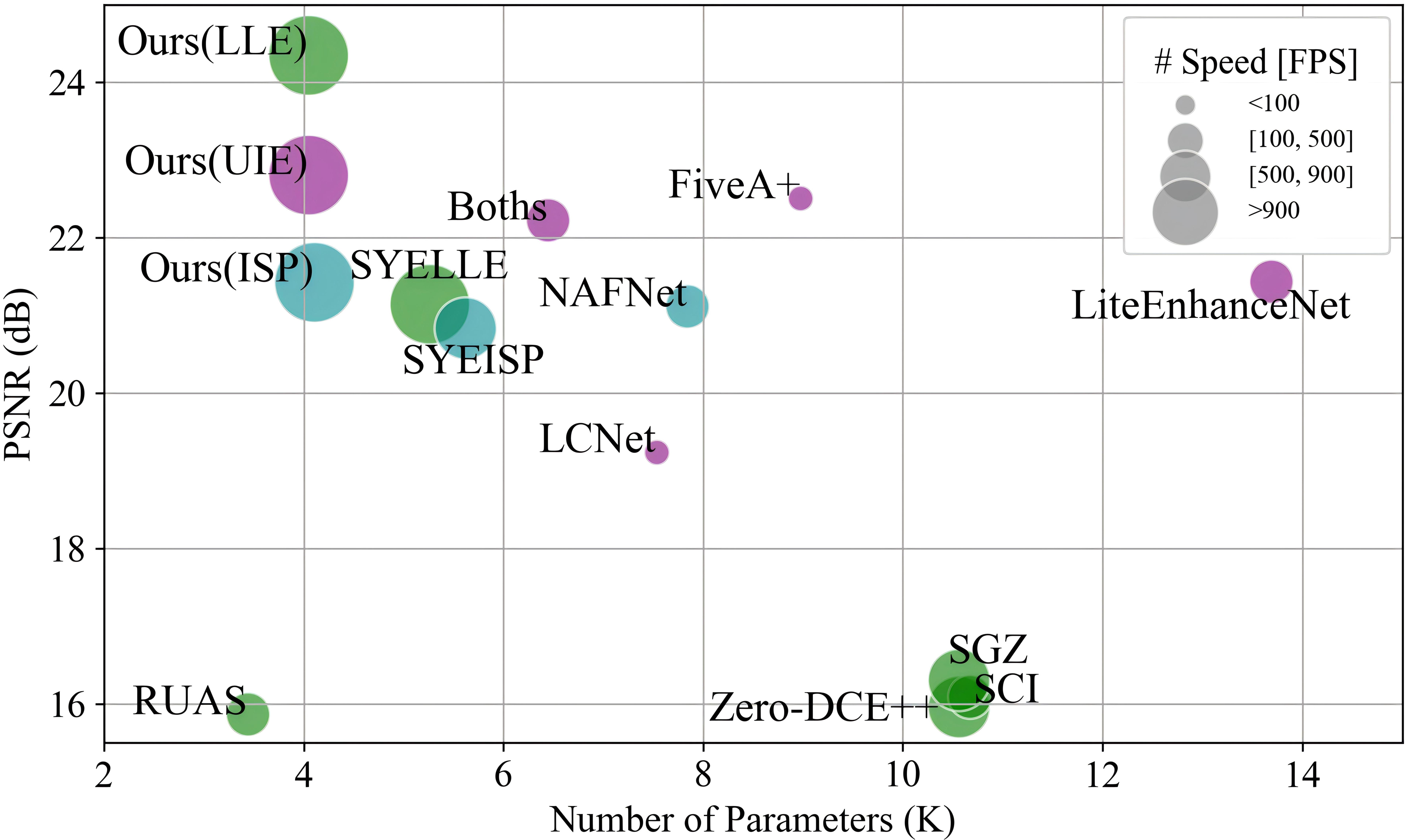} 
\caption{Efficiency comparison. MobileIE achieves superior performance in balancing speed, compactness, and accuracy.}
\label{fig1}
\vspace{-14pt}
\end{figure}

Recent advancements in Transformer-based \cite{choi2024reciprocal,li2023feature,wang2024correlation} and Diffusion-based \cite{xia2023diffir,zhao2024wavelet,yi2024diffraw} methods have demonstrated significant progress in image enhancement (IE) tasks. However, their reliance on computationally expensive self-attention mechanisms and iterative diffusion processes makes them unsuitable for mobile devices. While lightweight models have been developed to reduce computational complexity using parameter compression and low-FLOPs architecture designs \cite{reddy2023quantized, shi2023memory, ye2023accelir, qiao2024semi, xu2024boosting}, these approaches often compromise enhancement quality, making it challenging to achieve high performance across a wide range of degradation scenarios \cite{abrahamyan2021bias}. Furthermore, the increasing demand for high-resolution images \cite{wang2024correlation} exacerbates the computational burdens on mobile platforms.

Besides facing computational bottlenecks, current IE methods are typically optimized for specific degradation types, restricting their adaptability. Although advanced operators \cite{chen2024gcam,chen2023mofa} have shown effectiveness in certain tasks, their complexity makes them unsuitable for mobile deployment. Consequently, designing a universal and efficient network capable of real-time IE on resource-constrained devices remains a significant research challenge.


We believe that mobile-based IE should strike a balance between speed and performance, leveraging general architectures and hardware-friendly operators \cite{zhang2023rethinking,vasu2023mobileone}. To address these challenges, we introduce MobileIE, an efficient method designed for real-time enhancement in resource-constrained environments, optimizing both performance and resource utilization. Although degradation types vary widely, IE tasks share common requirements: global information for structural integrity and local details for accurate recovery. This commonality enables the use of universal modules across different tasks.

Building on this concept, MobileIE employs a streamlined topology and deployable operators to achieve efficient performance on resource-constrained devices. We decouple the training and inference phases, simplifying the feedforward structure for more efficient inference. At the core of our design is MBRConv, a Multi-Branch Re-parameterized Convolution, which captures multi-scale features using various convolution sizes. These features are then processed through concatenation, compression, and mapping to achieve the desired output dimensions. Additionally, we introduce a Feature Self-Transform (FST) module, which enhances the feature representation capability of linear convolutions by capturing nonlinear relationships through secondary feature interactions.

To enhance focus on critical regions, we simplify the attention mechanism and develop a hardware-friendly Hierarchical Dual-Path Attention (HDPA) mechanism, which efficiently fuses global and local features. To overcome training bottlenecks, we introduce an Incremental Weight Optimization (IWO) Strategy and a Local Variance Weighted (LVW) Loss function. The IWO strategy freezes prior knowledge, integrates it into trainable convolutional kernels, and re-parameterizes these into new kernels to boost model performance. Meanwhile, the LVW Loss function enhances accuracy without increasing the number of parameters. Key contributions of this paper include:

$\bullet$ We introduce MobileIE, a framework designed for real-time image enhancement on mobile devices, featuring compact and efficient modules tailored for resource-constrained environments.

$\bullet$ We propose an Incremental Weight Optimization (IWO) Strategy and a Local Variance Weighted (LVW) Loss to address the challenges of training compact models, enhancing performance without adding complexity.

$\bullet$ We demonstrate that MobileIE achieves state-of-the-art speed and performance across three image enhancement tasks, sustaining over 100 FPS and enabling seamless deployment on mobile devices.

\begin{figure*}[htbp]
\includegraphics[width=\linewidth]{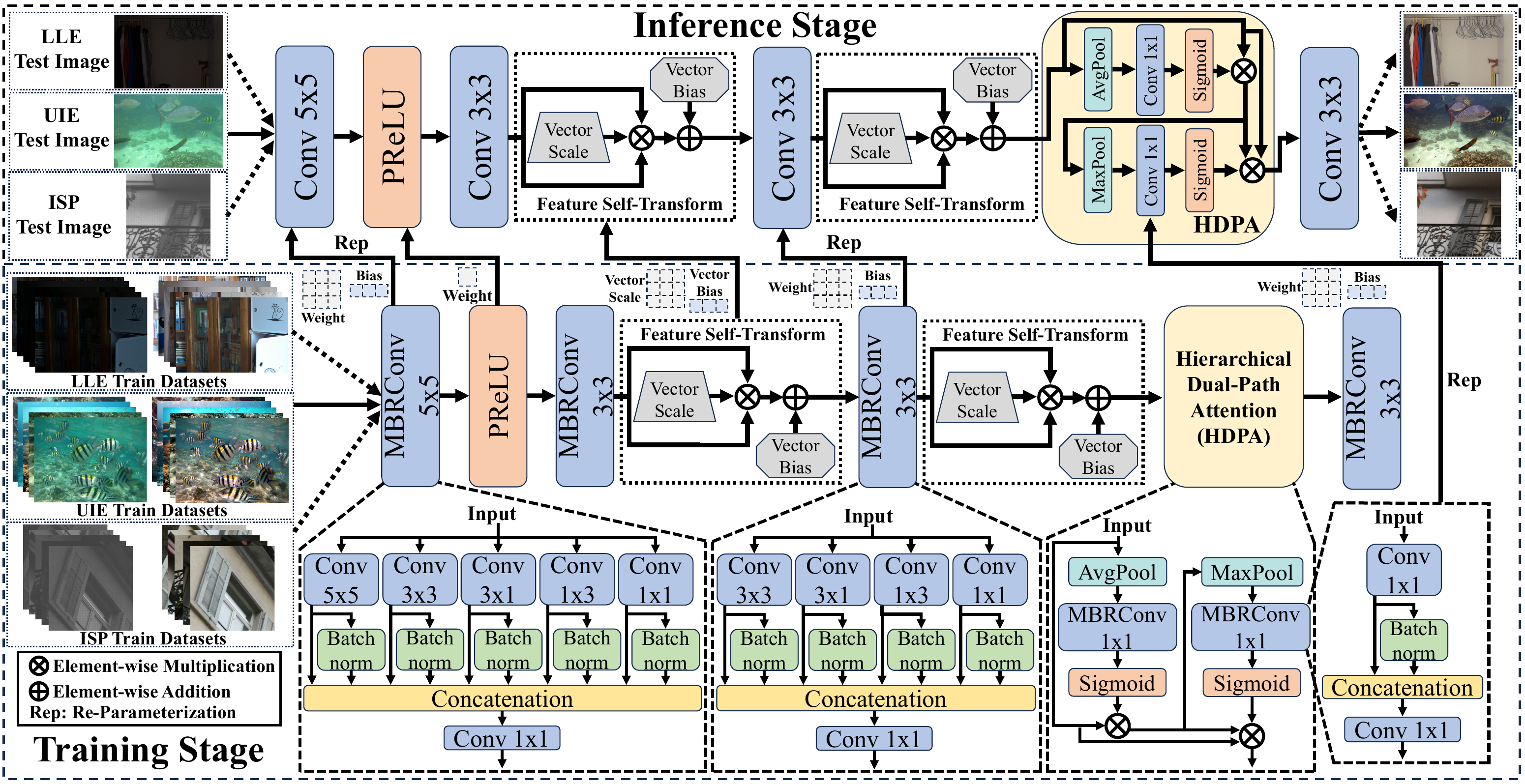}
\caption{\label{fig2}Architecture of the proposed MobileIE includes Multi-Branch Re-param Convolution (MBRConv), Feature Self-Transform (FST), and Hierarchical Dual-Path Attention (HDPA). During inference, MBRConv re-parameterizes into a standard convolution, reducing the model size to 4K parameters while maintaining training performance. The detailed architecture settings are provided in the Appendix.}
\vspace{-6pt}
\end{figure*}


\section{Related Works}
\subsection{Low-Level Vision Tasks} 
This section covers three key tasks: Low-Light Enhancement (LLE), Underwater Image Enhancement (UIE), and end-to-end Image Signal Processing (ISP).

\textbf{LLE.} LLE focuses on mitigating color distortion and noise in degraded images. Retinex-based \cite{wu2022uretinex, liu2021retinex,bai2024retinexmamba,yi2023diff,cai2023retinexformer} enhance images by separating illumination and reflectance. Zero-DCE \cite{guo2020zero} adjusts underexposed images using a luminance enhancement curve, and PairLIE \cite{fu2023learning} learns enhancement from varying lighting conditions. Transformer-based \cite{cai2023retinexformer,wen2024illumination,wang2023ultra} and Mamba-based \cite{bai2024retinexmamba,zhang2024llemamba} methods have also achieved notable success, while Diffusion-based \cite{yi2023diff, jiang2023low} methods excel at generating accurate images in extremely low-light conditions, even with noise and limited data.

\textbf{UIE.} Compared to traditional UIE methods \cite{zhuang2022underwater, drews2016underwater}, CNN-based \cite{islam2020fast, naik2021shallow, ma2022wavelet, sharma2023wavelength} directly learn feature representations from underwater images. Five A+ \cite{jiang2023five} proposed a two-stage framework with a pixel attention module for real-time enhancement. Transformer-based \cite{peng2023u,khan2024spectroformer} and Mamba-based \cite{an2024uwmamba,chen2024mambauie} models more effectively handle complex underwater scenes with their global receptive fields. Additionally, Diffusion-based \cite{zhao2024wavelet,tang2023underwater} approaches reduce noise and enhance contrast efficiently.

\textbf{ISP.} Learning-based ISP pipelines replace traditional multi-module systems by employing end-to-end models that directly process RAW input into RGB images. PyNet \cite{ignatov2020replacing} focuses on mobile ISPs, while MW-ISP \cite{ignatov2020aim} and AWNet \cite{dai2020awnet} improve performance using wavelet transforms. LiteISP \cite{zhang2021learning} resolves misalignment between RAW and sRGB images, achieving better results through joint learning. Rawformer \cite{perevozchikov2024rawformer} eliminates the need for paired datasets with an unsupervised Transformer, and DiffRAW \cite{yi2024diffraw} introduces diffusion models into ISP. Several ISP challenge works \cite{ignatov2020aim,ignatov2022learned,ignatov2021learned} have also produced notable results.
%
\subsection{Efficient Architectures Design}
Efficient neural networks aim to balance computational complexity and performance. Models like MobileNet \cite{howard2019searching} and ShuffleNet \cite{zhang2018shufflenet} achieve high performance with fewer FLOPs, while GhostNet \cite{han2020ghostnet} and FasterNet \cite{chen2023run} generate feature maps through cost-effective operations. StarNet \cite{ma2024rewrite} excels in extracting rich representations with star-shaped operations, and VanillaNet \cite{chen2024vanillanet} demonstrates minimalism's power using only a few convolutions.

Furthermore, re-parameterization strategies have also proven successful in efficient network design. Models like ACNet \cite{ding2019acnet}, DBB \cite{ding2021diverse}, and RepVGG \cite{ding2021repvgg} use multi-branch topologies to accelerate inference, while RepGhost \cite{ding2021repvgg}, RepViT \cite{wang2024repvit}, and MobileOne \cite{vasu2023mobileone} adopt similar approaches to enhance performance. Several NTIRE challenge submissions \cite{ren2024ninth,li2023ntire,conde2023efficient} have utilized re-parameterized structures to accelerate inference in efficient IE tasks.

\section{Proposed Method}
\subsection{Overall Pipeline}
We aim to design an efficient IE model that balances parameters, speed, and performance. Following the simplicity principle \cite{zhang2023rethinking}, we adopted a streamlined topology based on basic operations, illustrated in Figure \ref{fig2}.

The proposed MobileIE, built on Re-parameterization, integrates four key components: shallow feature extraction, deep feature extraction, feature transform, and an attention mechanism. During training, the degraded image first passes through MBRConv$5\times5$ and the PReLU activation function to extract shallow features. Two MBRConv$3\times3$ and FST modules then process these to learn deeper features. The HDPA mechanism directs the model’s focus to important regions, and MBRConv$3\times3$ refines the output. To address training performance bottlenecks, we introduce the IWO strategy and LVW loss function.

In inference, all MBRConv layers are re-parameterized into standard convolutions, reducing parameters while maintaining performance. MobileIE’s streamlined structure ensures fast inference and easy deployment on mobile devices. Despite its simplicity, MobileIE surpasses SOTA lightweight IE models, as shown in Figure \ref{fig1}.
\subsection{Multi-Branch Re-param Convolution}
Re-parameterization has achieved success in high-level vision tasks \cite{ding2019acnet,ding2021diverse,ding2021repvgg}, but yields unsatisfactory results when directly applied to image enhancement \cite{gou2023syenet}. We introduce MBRConv, specifically designed for image enhancement.

\begin{figure}[htbp]
\includegraphics[width=\linewidth]{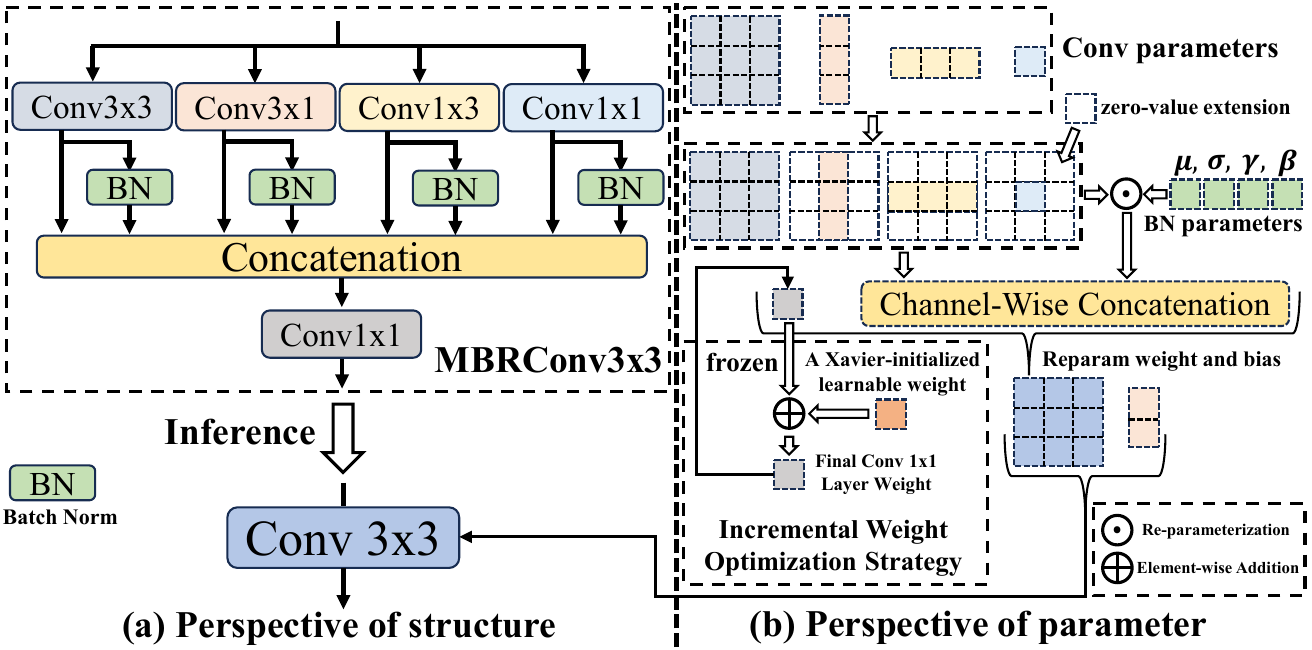}
\caption{\label{fig3}Structural of MBRConv$3\times3$.}
\end{figure}

Figure \ref{fig3}(a) shows MBRConv with multiple convolutional branches of varying kernel sizes, capturing multi-scale features that are concatenated and integrated via a Conv $1\times1$ ($conv_{out}$). The branches are re-parameterized during inference into a single convolution, reducing computational cost while preserving training performance. Unlike previous Rep methods, MBRConv includes parallel Batch Norm (BN) layers in each branch. Although BN is less effective for IE tasks \cite{wang2022repsr,zhang2021edge}, it enhances nonlinearity and cannot be replaced by other activations due to its unique ability to merge into a single convolution. The parallel BN layers retain both smoothed and original features, improving robustness across diverse data distributions. Additional details on MBRConv can be found in the Appendix.

Although multi-branch topologies assist in feature extraction, compact networks still struggle to learn complex features, leading to stagnation in later training stages. Adjusting training hyperparameters often results in marginal improvements. To address this, we propose the Incremental Weight Optimization (IWO) strategy, which combines learnable weights with prior knowledge to enhance the integration performance of the $conv_{out}$ by better capturing relationships between features across layers in MBRConv.

As shown in Figure \ref{fig3}(b), IWO combines two weight components: frozen weights, $W_{pre}$, from prior training, and learnable weights, $W_{learn}$. These are fused as:
\begin{equation}\label{(1)}
W_{final} = {\rm Frozen}(W_{pre}) + W_{learn},
\end{equation}
where $W_{pre}$ is the optimal weight from earlier training and remains frozen, and $W_{learn}$ is dynamically updated during subsequent training. The final weight, $W_{final}$, is applied to the concatenated feature map:
\begin{equation}\label{(1)}
y_{out} = F_{conv1\times1}(x_{concat};W_{final})+ b_{conv_{out}},
\end{equation}
where $x_{concat}$ denotes the concatenated multi-branch features, $F_{conv}$ represents the convolution. $y_{out}$ and $b_{conv_{out}}$ represent the output feature map and the bias, respectively. $W_{pre}$ offers a stable initial feature representation, minimizing redundancy. $W_{learn}$ refines these features, improving task-specific detail capture. IWO combines prior knowledge with learnable parameters, balancing knowledge transfer and feature refinement to improve feature integration.

\subsection{Feature Self-Transform}
While MBRConv captures multi-scale features, its linear operations restrict higher-order feature interactions. To address this, we introduce FST, which employs a quadratic interaction mechanism \cite{ma2024rewrite} to boost the model’s nonlinear expressiveness. In FST, the input features are element-wise multiplied by themselves, scaled by a learnable parameter, $Scale$, and adjusted by a learnable $bias$ term.
\begin{equation}\label{(1)}
{\rm FST}(x) = Scale\cdot(x\ast x) + bias.
\end{equation}
This quadratic interaction captures more complex relationships between features, unlike traditional linear combinations. A learnable $bias$ term is also applied channel-wise to fine-tune the output further. FST significantly improves feature representation by capturing higher-order interactions, while the learnable bias enhances adaptability across dimensions. This design balances improved expressiveness with computational efficiency, making FST suitable for lightweight models and real-time inference. 


\subsection{Hierarchical Dual-Path Attention}
We propose a simple HDPA mechanism to improve feature extraction in IE. HDPA selectively captures both global and local features, integrating spatial and channel interactions to improve feature precision and comprehensiveness. Its dual-path structure optimizes feature selection at each layer through two distinct pathways. HDPA operates in two steps:

(1) Global Feature Extraction: Global statistics are extracted using adaptive average pooling. Global features are processed by an MBRConv1x1 layer, and Sigmoid activation to generate channel-wise attention weights, modulating the importance of each channel. This step is expressed as:
\begin{equation}\label{(1)}
A_{g} = {\rm Sigmoid(MBRConv1\times1(AvgPool}(F))),
\end{equation}
where $A_{g}$ denotes the global attention weights, and $F$ represents the input features.

(2) Local Feature Enhancement: In the second pathway, local features are extracted using max pooling on globally enhanced features. The input is multiplied by the global weights $W_{g}=F{\ast}A_{g}$, followed by max pooling to capture local information. The local responses are processed by an MBRConv$1\times1$, with Sigmoid activation generating the local attention weights.
\begin{equation}\label{(1)}
A_{l} = {\rm Sigmoid(MBRConv1\times1(MaxPool}(W_{g}))),
\end{equation}
where $A_{l}$ represents the local attention weights.

Finally, the global and local attention weights are combined through element-wise multiplication to produce the final attention feature, which is applied to the input features:
\begin{equation}\label{(1)}
\hat{F} = {\rm HDPA}(F)=(A_{g}\ast A_{l}) \ast F,  
\end{equation}
where $\hat{F}$ donates the output feature after HDPA processing.

This dual-path hierarchical design balances global context and local feature enhancement, improving both depth and accuracy in feature extraction. Its simplicity reduces the computational cost, making it well-suited for resource-constrained environments.
\subsection{Local Variance Weighted Loss}

Compact CNNs, with limited parameters, struggle to capture rich features and are sensitive to extreme pixels that hinder optimization \cite{abrahamyan2021bias, gou2023syenet}. While $L1$ loss is robust, its equal weighting limits outlier handling; $L2$ emphasizes outliers but may cause imbalance. To address this, we introduce Local Variance Weighted Loss (LVW), which adapts to local variability and mitigates outliers effectively.

\begin{figure}[htbp]
\begin{minipage}[htbp]{0.5\linewidth}
\centerline{\includegraphics[width=\linewidth]{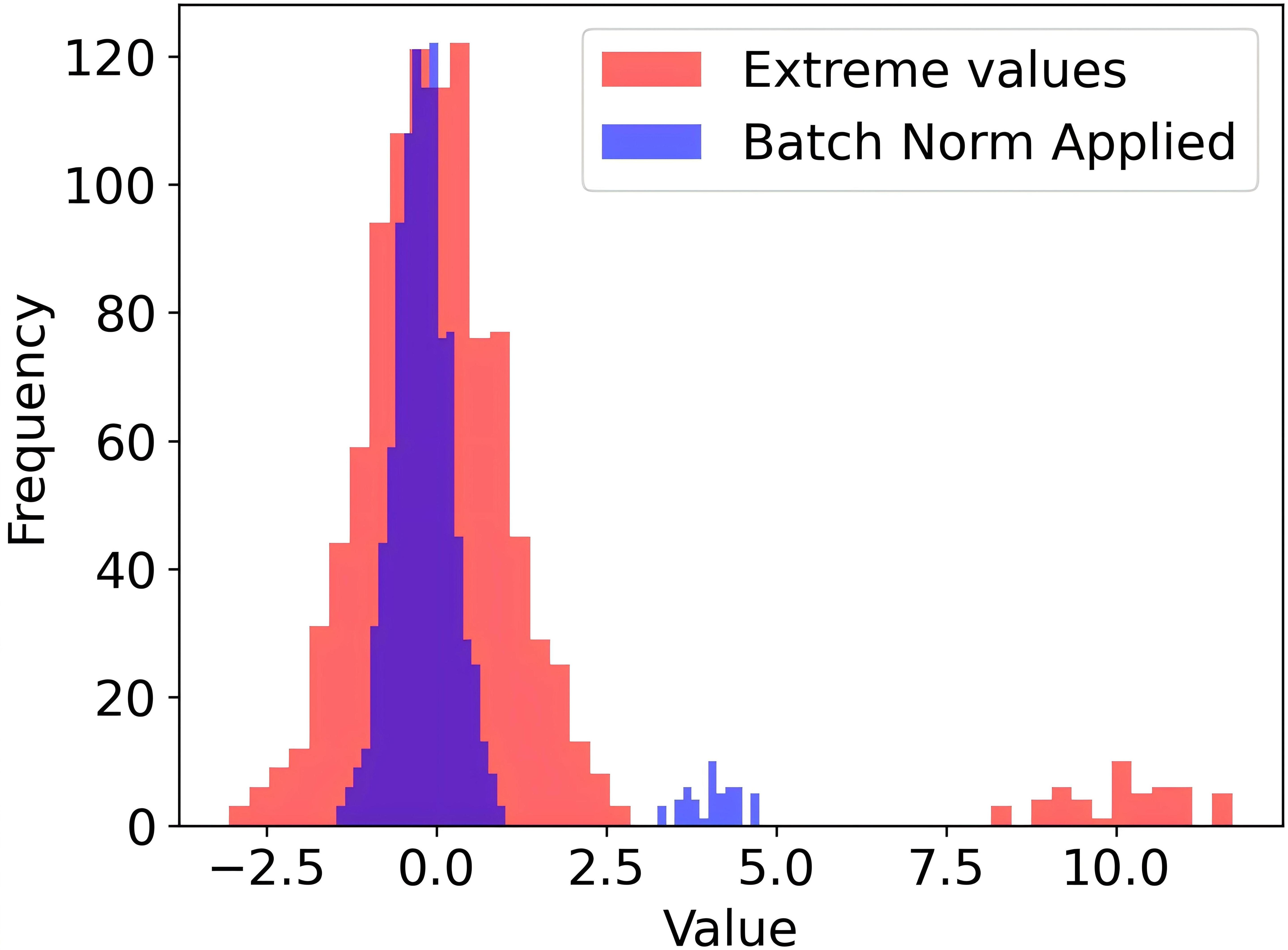}}
\centerline{\scriptsize\textbf{(a)}}
\end{minipage}%
\begin{minipage}[htbp]{0.5\linewidth}
\centerline{\includegraphics[width=\linewidth]{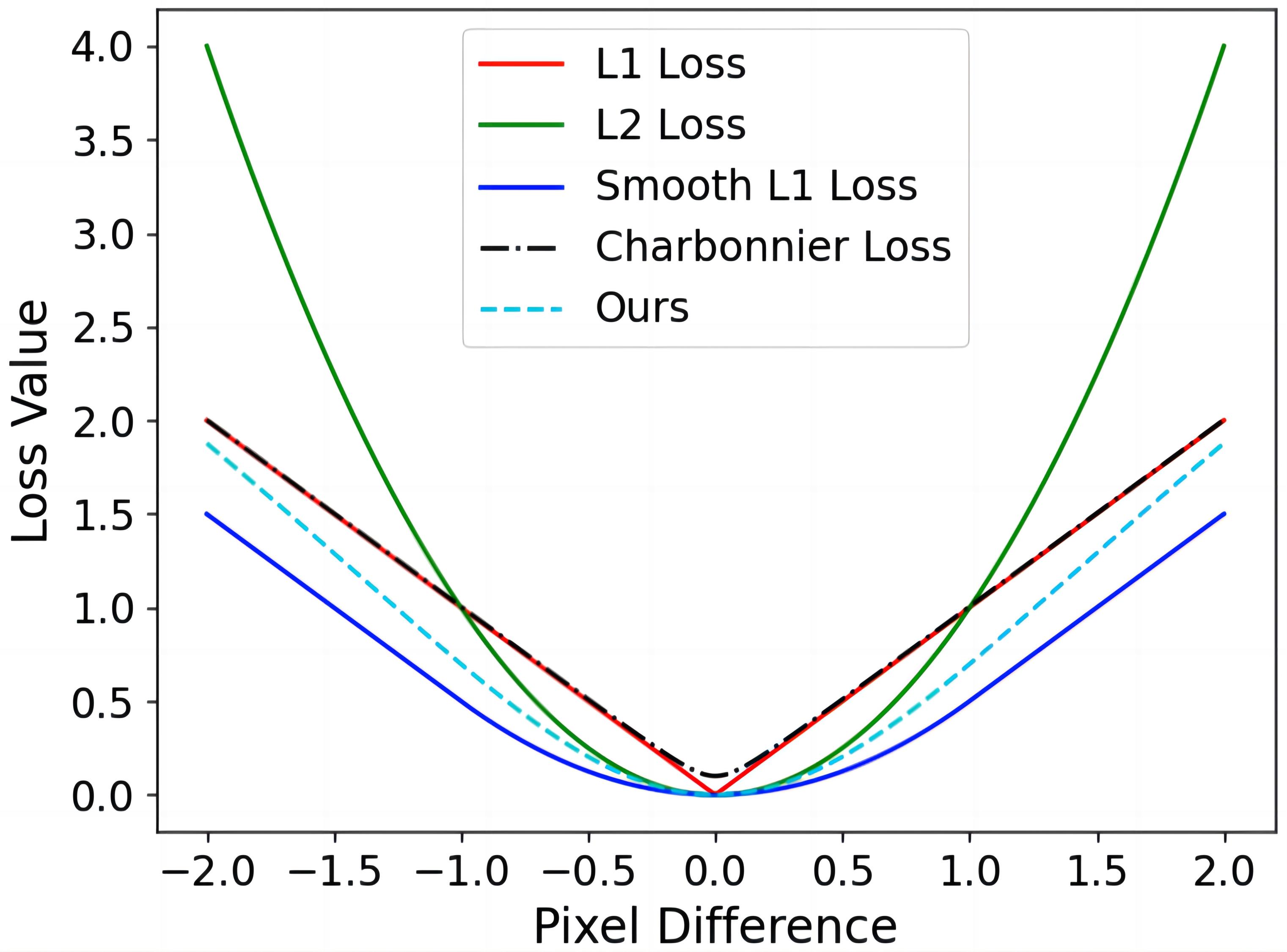}}
\centerline{\scriptsize\textbf{(b)}}
\end{minipage}%
\caption{\label{loss}(a) Effect of Batch Norm on Data Distribution. (b) Comparison of Loss Functions for Pixel Difference Handling.}
\end{figure}
\begin{table*}[htbp]
  \centering
  \resizebox{\textwidth}{!}{
    \begin{tabular}{ccccccc|cccc|cccc}                 
    \toprule
    \multicolumn{1}{c}{Method} & \multicolumn{1}{c}{Venue} & \multicolumn{1}{c}{\#Params$\downarrow$} & \multicolumn{1}{c}{Model Size$\downarrow$} & \multicolumn{1}{c}{Latency$\downarrow$} & \multicolumn{1}{c}{Latency$\downarrow$} & \multicolumn{1}{c|}{FPS$\uparrow$} & \multicolumn{4}{c|}{LOLv1 \cite{wei2018deep}} & \multicolumn{4}{c}{LOLv2-Real \cite{yang2020fidelity}} \\
    \cmidrule{8-15}
    & & (K) & (MB) & (GPU,ms) & (SoC,ms) & (600$\times$400) & PSNR$\uparrow$ & SSIM$\uparrow$ & LPIPS$\downarrow$ &SCORE$\uparrow$& PSNR$\uparrow$ & SSIM$\uparrow$ & LPIPS$\downarrow$ & SCORE$\uparrow$ \\
    \hline
    Kind++ \cite{zhang2021beyond} & IJCV'21 & 8,280 & / & \textgreater 500 & \textgreater 500 & \multicolumn{1}{c|}{\textless 10} & 17.75 & 0.766 & 0.198 & \textless 0.001 & 17.66 & 0.770 & 0.217 & \textless 0.001  \\
    DDNet \cite{qu2024double} & IEEE TITS'24 & 5,389 & 20.56 & 32.910 & \textgreater 500 & 30.792 & 21.82 & 0.802 & \textcolor{red}{0.186} & 0.208 & 23.02 & 0.838 & \textcolor{red}{0.173} & 1.099 \\
    PairLIE \cite{fu2023learning} & CVPR'23 & 341.767 & 1.30 & 11.580 & \textgreater 500 & 86.354 & 19.51 & 0.736 & 0.248 & 0.024 & 19.88 & 0.778 & 0.234 & 0.040\\
    IAT \cite{cui2022you} & BMVC'22 & 86.856 & 0.33 & 6.204 & 202.33 & 161.186 & \textcolor{blue}{23.38} & 0.808 & 0.216 & 9.604 & \textcolor{red}{25.46} & 0.843 & 0.182 & \textcolor{blue}{171.690}\\
    Zero-DCE \cite{guo2020zero} & CVPR'20 & 79.416 & 0.30 & 2.539 & 82.94 & 393.910 & 14.86 & 0.559 & 0.335 & \textless 0.001 & 18.06 & 0.574 & 0.313 & 0.015 \\
    3DLUT \cite{zeng2020learning} & IEEE TPAMI'20 & 593.500 & 2.26 & 1.176 & / & 850.340 & 17.59 & 0.721 & 0.232 & 0.218 & 19.68 & 0.637 & 0.224 & 6.000\\
    Zero-DCE++ \cite{li2021learning} & IEEE TPAMI'21 & 10.561 & 0.04 & 1.974 & 57.91 & 506.558 & 14.68 & 0.472 & 0.340 & \textless 0.001 & 17.23 & 0.412 & 0.319 & 0.006 \\
    SCI \cite{ma2022toward} & CVPR'22 & 10.671 & 0.04 & 4.773 & 116.8 & 209.525 & 14.90 & 0.531 & 0.341 & \textless 0.001 & 17.30 & 0.534 & 0.308 & 0.003 \\
    SGZ \cite{zheng2022semantic} & WACV'22 & 10.561 & 0.04 & 1.970 & 59.27 & 507.662 & 15.28 & 0.473 & 0.339 & \textless 0.001 & 17.34 & 0.409 & 0.322 & 0.007\\
    RUAS \cite{liu2021retinex} & CVPR'21 & \textcolor{red}{3.438} & \textcolor{red}{0.01} & 4.421 & 109.15 & 226.175 & 16.40 & 0.500 & 0.270 & \textless 0.001 & 15.33 & 0.488 & 0.310 & \textless 0.001\\
    SYELLE \cite{gou2023syenet} & ICCV'23 & 5.259 & \textcolor{blue}{0.02} & \textcolor{blue}{0.944} & \textcolor{blue}{7.73} & \textcolor{blue}{1059.732} & 21.03 & 0.794 & 0.219 & 2.428 & 21.26 & 0.801 & 0.308 & 3.340 \\
    Adv-LIE \cite{wang2024adversarially} & MMM'24 & 238.560 & 0.91 & 14.269 & \textgreater 500  & 70.083 & 23.02 & \textcolor{blue}{0.808} & 0.203 & \textcolor{blue}{2.535} & 21.95 & \textcolor{blue}{0.844} & 0.192 & 0.575 \\
    \hline
    Ours & / & \textcolor{blue}{4.047} & \textcolor{blue}{0.02} & \textcolor{red}{0.895} & \textcolor{red}{6.72} & \textcolor{red}{1120.584} & \textcolor{red}{23.62} & \textcolor{red}{0.812} & \textcolor{blue}{0.198} & \textcolor{red}{92.855} & \textcolor{blue}{25.08} & \textcolor{red}{0.845} & \textcolor{blue}{0.180} & \textcolor{red}{702.767}\\
    \bottomrule
    \end{tabular}}
  \caption{Performance comparison of different low-light image enhancement models on LOL datasets. SCORE \cite{ignatov2022learned} represents a comprehensive measure of model performance and efficiency. The top results are marked: best in red and second in blue.}
  \label{tab:1}
\end{table*}

Let \{$\textbf{O}$, $\textbf{L}\}\in{\mathbb{R}^{B{\times}C{\times}H{\times}W}}$ denote the predicted output and ground-truth, where $B, C, H$, and $W$ denote batch size, number of channels, height, and width of the image, respectively. The absolute difference between the predicted output and the ground truth at pixel $(m,n)$ is calculated as:
\begin{equation}\label{(1)}
\Delta_{m,n} = \Arrowvert{O_{m,n} - L_{m,n}\Arrowvert}_{1}.
\end{equation}

For each predicted output, we calculate the local mean $\mu_{m, n}$ and variance $\sigma^{2}_{m, n}$ of $\Delta_{m,n}$ across the spatial dimensions $H$ and $W$. This can be expressed as follows:
\begin{equation}\label{(5)}
\left\{\begin{aligned}
&\mu_{m, n} = \frac{1}{H\cdot W}(\sum^{H}_{m=1}\sum^{W}_{n=1}\Delta_{m,n})\\
&\sigma^{2}_{m, n} = \frac{1}{H\cdot W}(\sum^{H}_{m=1}\sum^{W}_{n=1}\Delta_{m,n}-\mu_{m, n})^{2}\\
	\end{aligned}
	\right
	.
\end{equation}

To demonstrate the impact of mean and variance on outliers, we applied BN to simulated data with both normal values and outliers (Figure \ref{loss}(a)). BN balances sample contributions to mean and variance, reducing outliers' influence on the distribution. Based on these local statistics, we compute a weighted factor for each pixel. The weight $W_{\Delta}$, is normalized by the absolute deviation of the error relative to the local mean and scaled according to the local variance:
\begin{equation}\label{(1)}
W_{\Delta} = {\rm Tanh}( \frac{|\Delta_{m,n}-\mu_{m,n}|}{\sigma_{m, n}+\varepsilon}),
\end{equation}
where $\varepsilon$ is a small constant added to prevent division by zero. The Tanh function is applied to ensure the weights remain within a bounded range, facilitating a smooth transition between high-variance and low-variance regions.

The final loss is calculated by multiplying $\Delta_{m,n}$ with the corresponding local weighted factor:
\begin{equation}\label{(1)}
\mathcal{L}_{LVW}=\frac{1}{H\cdot W}\sum^{H}_{m=1}\sum^{W}_{n=1}(W_{\Delta}\cdot\Delta_{m,n}).
\end{equation}

\begin{table*}[htbp]
\centering
  \resizebox{0.9\textwidth}{!}{
    \begin{tabular}{ccccccc|cccc}
    \toprule
    \multicolumn{1}{c}{Method} & \multicolumn{1}{c}{Venue} & \multicolumn{1}{c}{\#Params$\downarrow$} & \multicolumn{1}{c}{Model Size$\downarrow$} & \multicolumn{1}{c}{Latency$\downarrow$} & \multicolumn{1}{c}{Latency$\downarrow$} & \multicolumn{1}{c|}{FPS$\uparrow$} & \multicolumn{4}{c}{UIEB \cite{li2019underwater}} \\
    \cmidrule{8-11}
    & & (K) & (MB) & (GPU,ms) & (SoC,ms) & (640$\times$480) & PSNR$\uparrow$ & SSIM$\uparrow$ & LPIPS$\downarrow$ & SCORE$\uparrow$\\
    \hline 
    FUnIE-GAN \cite{islam2020fast} & IEEE RA-L'20 & 7,020 & 26.78 & 3.698 & 76.61 & 270.428 & 19.72 & 0.845 & 0.239 & 0.101 \\
    Shallow-UWNet \cite{naik2021shallow} & AAAI'21 & 219.456 & 0.84 & 11.158 & 400.01 & 89.620 & 16.69 & 0.747 & 0.365 & \textless 0.001\\
    PUIE \cite{fu2022uncertainty}  & ECCV'22 & 1,401 & 5.34 & 24.000 & \textgreater 500 & 41.658 & 21.25 & 0.885 & 0.161 & 0.130 \\
    UIE-WD \cite{ma2022wavelet} & ICASSP'22 & 13,704 & 52.28 & 10.235 & 386.72 & 97.920 & 20.92 & 0.847 & 0.212 & 0.192 \\
    U-Shape \cite{peng2023u} & IEEE TIP'23 & 22,817 & 87.04 & 38.484 & \textgreater 500 & 25.984 & 21.25 & 0.845 & 0.198 & 0.081 \\
    FiveA+ \cite{jiang2023five} & BMVC'23 & 8.974 & 0.03 & 11.700 & 423.43 & 90.224 & \textcolor{blue}{22.51} & 0.902 & 0.165 & 1.525\\
    Boths \cite{liu2022boths} & IEEE GRSL'23 & \textcolor{blue}{6.447} & \textcolor{red}{0.02} & \textcolor{blue}{2.567} & \textcolor{blue}{58.04} & \textcolor{blue}{389.431} & 22.23 & \textcolor{blue}{0.904} & \textcolor{blue}{0.156} & \textcolor{blue}{2.173} \\
    SFGNet \cite{zhao2024toward} & ICASSP'24 & 1,298 & 4.95 & \textgreater 100 & \textgreater 500 & \textless 10 & 21.66 & 0.871 & 0.191 & \textless 0.001 \\
    LiteEnhanceNet \cite{zhang2024liteenhancenet} & ESWA'24 & 13.688 & 0.05 & 6.654 & 190.02 & 150.289 & 21.44 & 0.903 & 0.168 & 0.608\\
    LSNet \cite{zhou20247k} & Arxiv'24 & 7.534 & \textcolor{blue}{0.03} & 59.519 & \textgreater 500 & 16.810 & 19.24 & 0.829 & 0.242 & 0.003\\
    \hline
    Ours & / & \textcolor{red}{4.047} & \textcolor{red}{0.02} & \textcolor{red}{0.910} & \textcolor{red}{8.94} & \textcolor{red}{1099.370} & \textcolor{red}{22.81} & \textcolor{red}{0.906} & \textcolor{red}{0.155} & \textcolor{red}{29.711}\\
    \bottomrule
    \end{tabular}}
\caption{Performance comparison of different underwater image enhancement models on UIEB datasets. SCORE \cite{ignatov2022learned} represents a comprehensive measure of model performance and efficiency. The top results are marked: best in red and second in blue.}
  \label{tab:2}
\end{table*}

$\mathcal{L}_{LVW}$ dynamically adjusts each pixel's contribution to the overall loss based on locally computed weights, thereby placing greater emphasis on optimizing anomalous pixels.

Figure \ref{loss}(b) illustrates how different loss functions handle pixel differences, emphasizing their handling of prediction errors. $\mathcal{L}_{LVW}$ avoids the steep increase of $L2$ loss for large errors, reducing sensitivity to extreme pixels while preserving enough gradient response for small errors, balancing robustness and detail preservation.

\section{Experiments}
\subsection{Experimental Settings}
\textbf{Implementation Details.} We implemented MobileIE in PyTorch. The model uses the Adam optimizer with a cosine annealing learning rate schedule, starting at 0.001. The learning rate is reset every 50 epochs with gradual decay. A 10-epoch warm-up phase \cite{he2022masked} is applied with a fixed learning rate of 1e-6. The model is trained for 2,000 epochs, incorporating an Incremental Weight Optimization Strategy for improved convergence. For the ISP task, input data is preprocessed into a $256 \times 256$ Bayer pattern.

\textbf{Dataset and Metrics.} For LLE task, the LOLv1 \cite{wei2018deep} and LOLv2 \cite{yang2020fidelity} datasets are used for both training and testing. For UIE and ISP tasks, the UIEB \cite{li2019underwater} and ZRR \cite{ignatov2020replacing} datasets are utilized, respectively. 

\subsection{Quantitative and Visual Comparisons}
In this section, we compare the proposed MobileIE with current lightweight SOTA methods across three IE tasks, focusing on visual comparisons and performance metrics. Specifically, PSNR, SSIM, LPIPS, and SCORE \cite{ignatov2022learned} are used as evaluation metrics, while computational complexity is tested on a single NVIDIA 4090 (GPU) and a smartphone with a Snapdragon 8 Gen 3 System on Chip (SoC).

\begin{figure}[b]
\centering
\vspace{-12pt}
\begin{minipage}[htbp]{0.242\linewidth}
\centerline{\includegraphics[width=\textwidth]{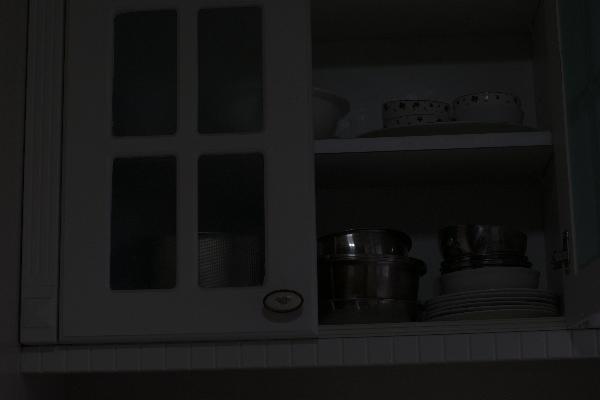}}
\centerline{\scriptsize\textbf{(a) Input}}
\centerline{\includegraphics[width=\textwidth]{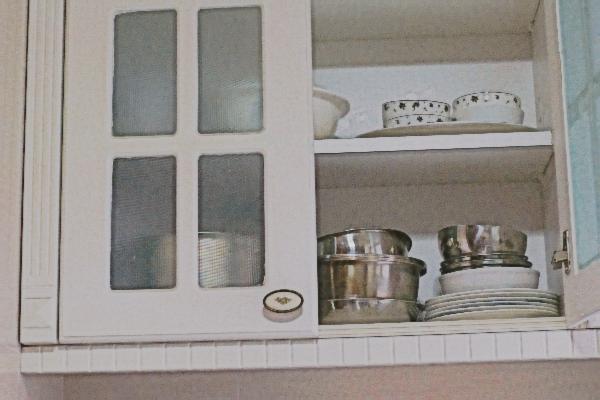}}
\centerline{\scriptsize\textbf{(f) PairLIE~\cite{fu2023learning}}}
\end{minipage}%
\hspace{0.0000000001\linewidth} 
\begin{minipage}[htbp]{0.242\linewidth}
\centerline{\includegraphics[width=\textwidth]{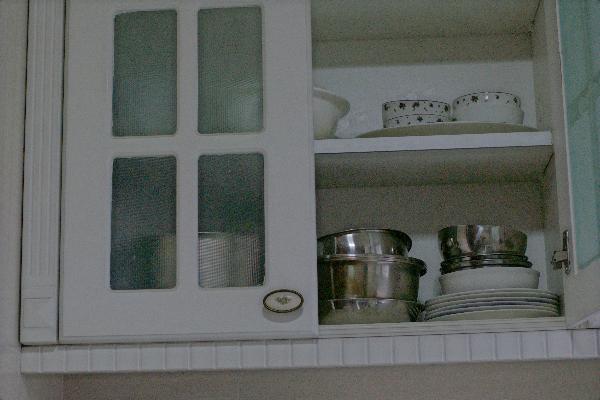}}
\centerline{\scriptsize\textbf{(b) Zero-DCE~\cite{guo2020zero} }}
\centerline{\includegraphics[width=\textwidth]{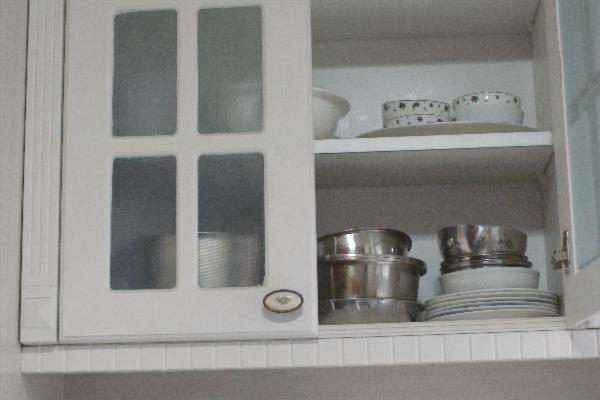}}
\centerline{\scriptsize\textbf{(e) SYELLE~\cite{gou2023syenet}}}
\end{minipage}%
\hspace{0.0000000001\linewidth}  
\begin{minipage}[htbp]{0.242\linewidth}
\centerline{\includegraphics[width=\textwidth]{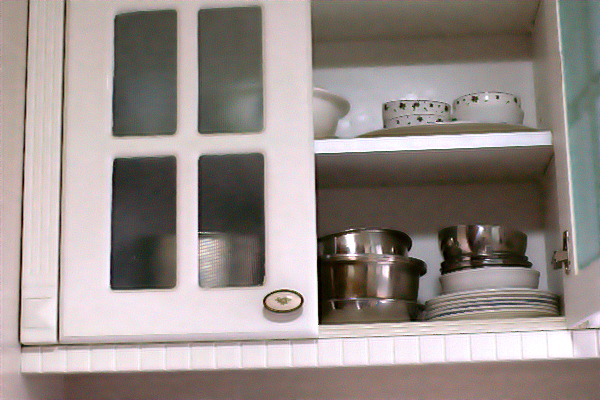}}
\centerline{\scriptsize\textbf{(c) RUAS~\cite{liu2021retinex}}}
\centerline{\includegraphics[width=\textwidth]{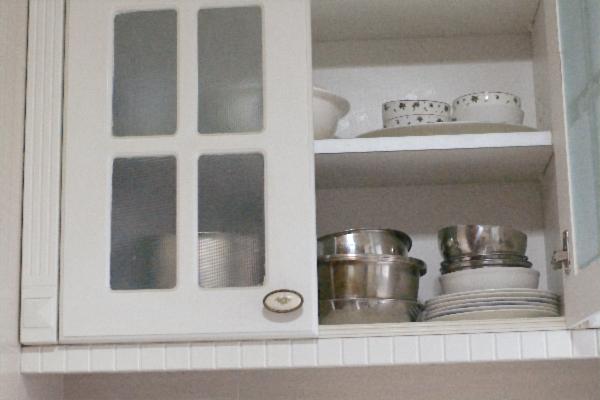}}
\centerline{\scriptsize\textbf{(i) Ours}}
\end{minipage}%
\hspace{0.0000000001\linewidth} 
\begin{minipage}[htbp]{0.242\linewidth}
\centerline{\includegraphics[width=\textwidth]{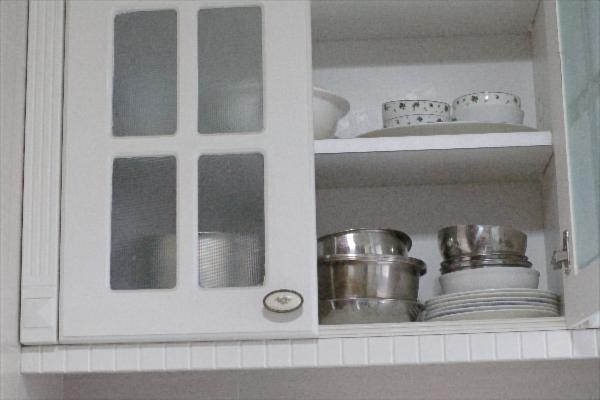}}
\centerline{\scriptsize\textbf{(h) IAT~\cite{cui2022you}}}
\centerline{\includegraphics[width=\textwidth]{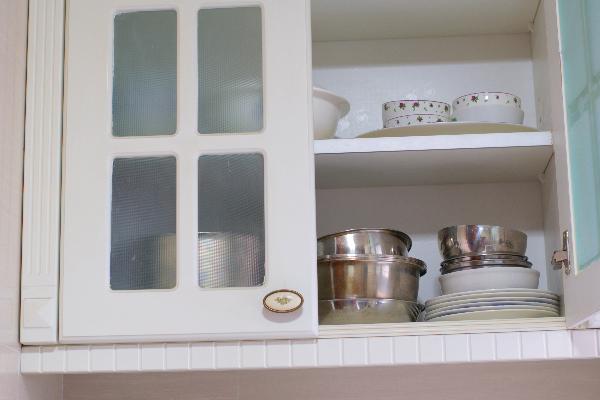}}
\centerline{\scriptsize\textbf{(j) Ground Truth}}
\end{minipage}%
\vspace{-6pt}
\caption{\label{lle} Visualization comparison of LLE on LOLv1 \cite{wei2018deep}.}
\end{figure}

\textbf{LLE:} The quantitative results on the LOLv1 \cite{wei2018deep} and LOLv2 \cite{yang2020fidelity} datasets are presented in Table \ref{tab:1}. MobileIE achieves superior or comparable PSNR and SSIM scores across all metrics, with an inference speed of \textbf{0.895 ms}. Compared to the lightweight SOTA method IAT \cite{cui2022you}, MobileIE delivers comparable PSNR gains while using only \textbf{4.7\%} of the parameters, with a \textbf{6.9x} faster inference speed. Figure \ref{lle} shows the visual results, where MobileIE produces higher-quality images, closely resembling the ground truth.

\begin{figure}[b]
\centering
\vspace{-6pt}
\begin{minipage}[htbp]{0.242\linewidth}
\centerline{\includegraphics[width=\textwidth]{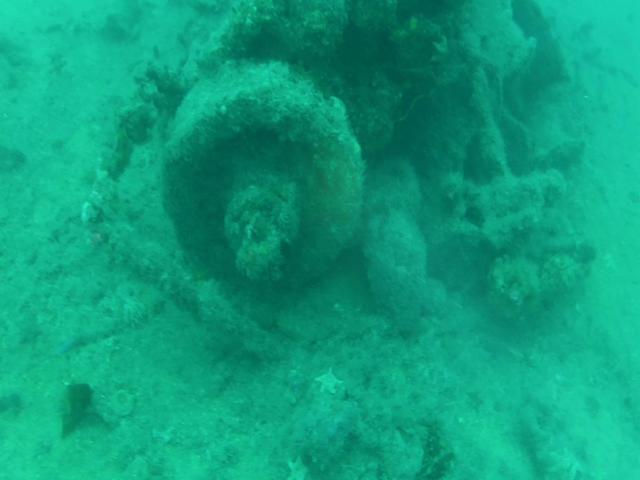}}
\centerline{\scriptsize\textbf{(a) Input}}
\centerline{\includegraphics[width=\textwidth]{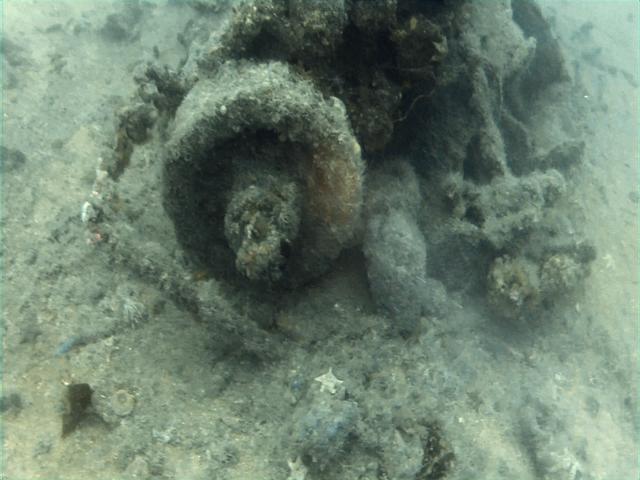}}
\centerline{\scriptsize\textbf{(e) LiteEnhanceNet~\cite{zhang2024liteenhancenet}}}
\end{minipage}%
\hspace{0.000001\linewidth} 
\begin{minipage}[htbp]{0.242\linewidth}
\centerline{\includegraphics[width=\textwidth]{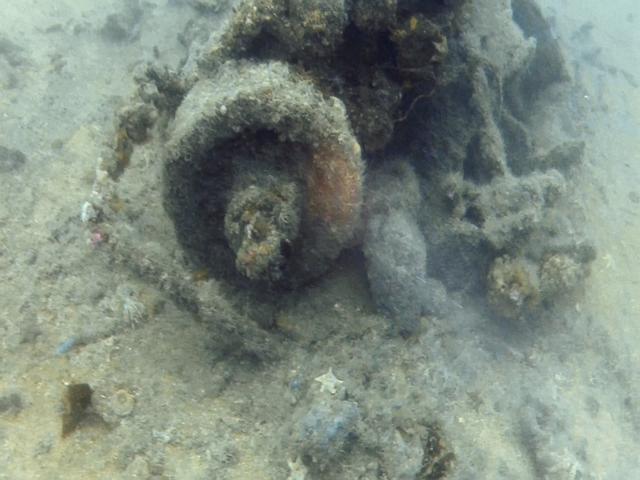}}
\centerline{\scriptsize\textbf{(b) PUIE~\cite{fu2022uncertainty}}}
\centerline{\includegraphics[width=\textwidth]{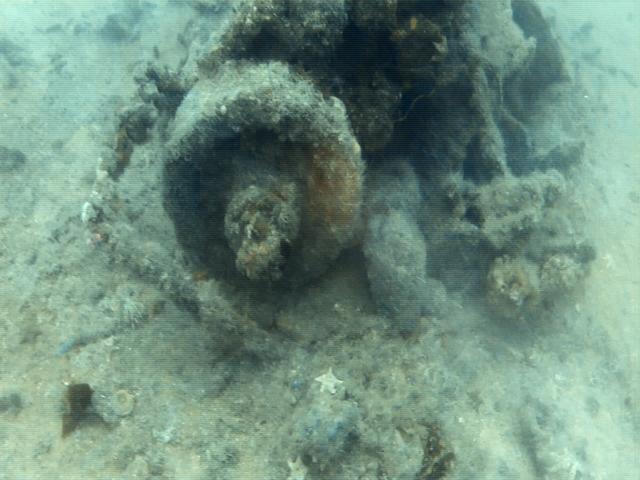}}
\centerline{\scriptsize\textbf{(f) LSNet~\cite{zhou20247k}}}
\end{minipage}%
\hspace{0.000001\linewidth} 
\begin{minipage}[htbp]{0.242\linewidth}
\centerline{\includegraphics[width=\textwidth]{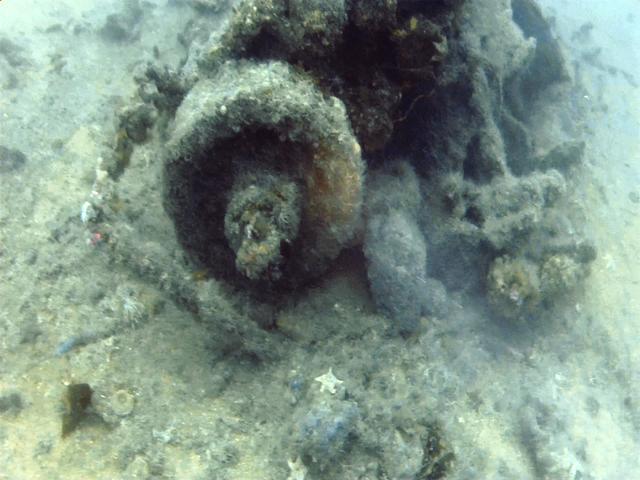}}
\centerline{\scriptsize\textbf{(c) FivaA+~\cite{jiang2023five}}}
\centerline{\includegraphics[width=\textwidth]{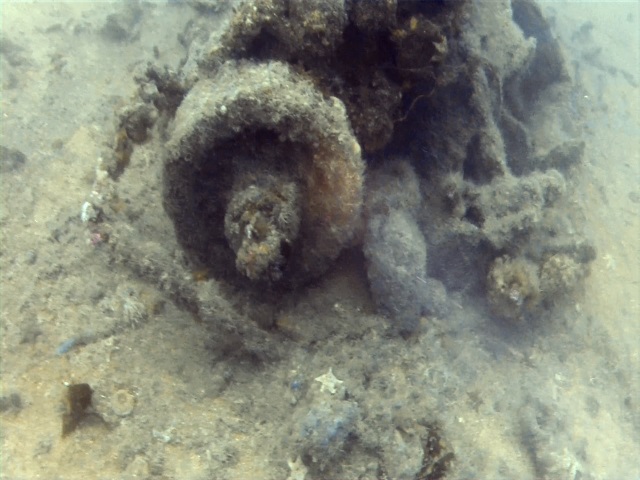}}
\centerline{\scriptsize\textbf{(g) Ours}}
\end{minipage}%
\hspace{0.000001\linewidth} 
\begin{minipage}[htbp]{0.242\linewidth}
\centerline{\includegraphics[width=\textwidth]{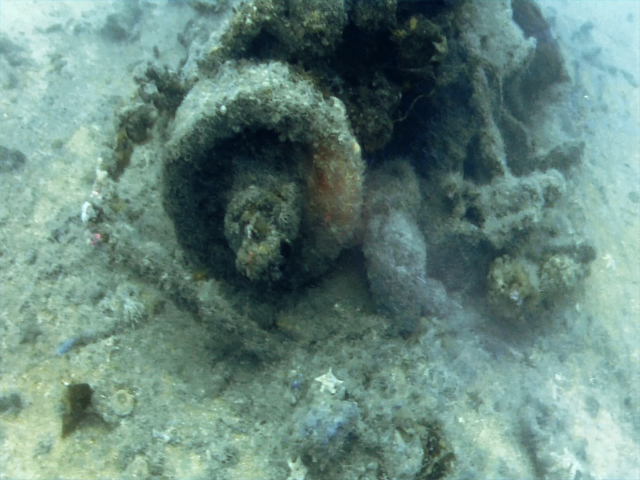}}
\centerline{\scriptsize\textbf{(d) Boths~\cite{liu2022boths}}}
\centerline{\includegraphics[width=\textwidth]{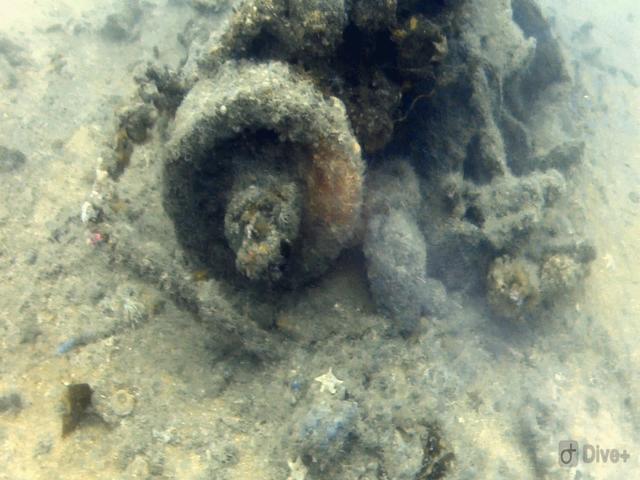}}
\centerline{\scriptsize\textbf{(h) Ground Truth}}
\end{minipage}%
\vspace{-6pt}
\caption{\label{uie}Visualization comparison of UIE on UIEB \cite{li2019underwater}.}
\end{figure}

\textbf{UIE:} The quantitative results in Table \ref{tab:2} show that MobileIE achieves a PSNR of 22.81 dB. Remarkably, MobileIE surpasses FiveA+ \cite{jiang2023five} by 0.3 dB in PSNR while using only \textbf{$45\%$} of its parameters. Figure \ref{uie} provides visual comparisons, highlighting MobileIE's superior color restoration compared to other lightweight methods.

\begin{figure}[htbp]
\centering
\begin{minipage}[htbp]{0.226\linewidth}  
\centerline{\includegraphics[width=\textwidth]{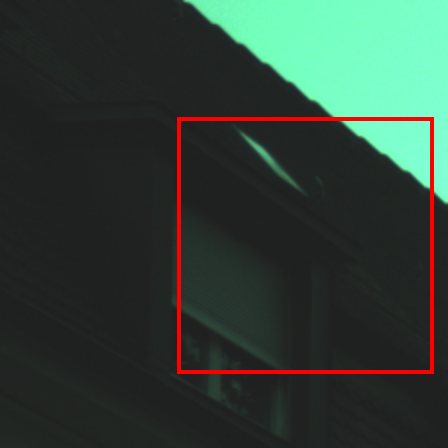}}
\centerline{\scriptsize\textbf{(a) Raw visualized}}
\centerline{\includegraphics[width=\textwidth]{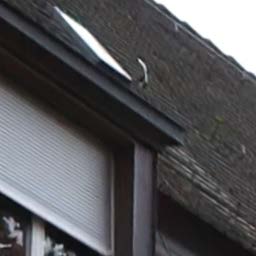}}
\centerline{\scriptsize\textbf{(e) LiteISPNet \cite{zhang2021learning}}}
\end{minipage}%
\hspace{0.001\linewidth} 
\begin{minipage}[htbp]{0.226\linewidth}  
\centerline{\includegraphics[width=\textwidth]{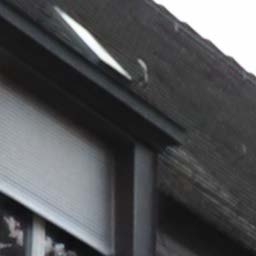}}
\centerline{\scriptsize\textbf{(b) PyNet \cite{ignatov2020replacing}}}
\centerline{\includegraphics[width=\textwidth]{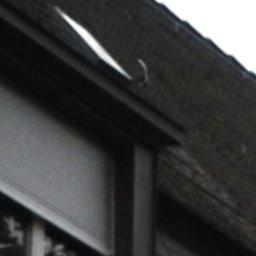}}
\centerline{\scriptsize\textbf{(f) SYEISP \cite{gou2023syenet}}}
\end{minipage}%
\hspace{0.001\linewidth} 
\begin{minipage}[htbp]{0.226\linewidth}
\centerline{\includegraphics[width=\textwidth]{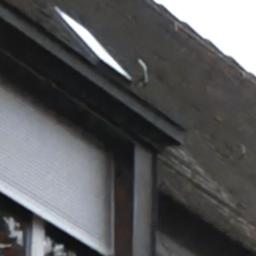}}
\centerline{\scriptsize\textbf{(c) AWNet \cite{dai2020awnet}}}
\centerline{\includegraphics[width=\textwidth]{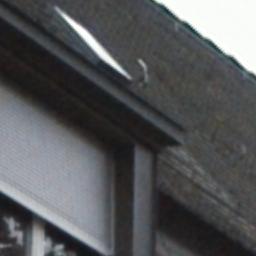}}
\centerline{\scriptsize\textbf{(g) Ours}}
\end{minipage}%
\hspace{0.001\linewidth} 
\begin{minipage}[htbp]{0.226\linewidth}
\centerline{\includegraphics[width=\textwidth]{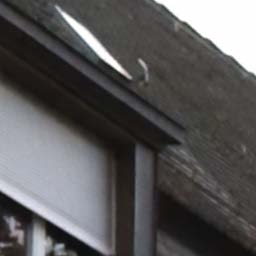}}
\centerline{\scriptsize\textbf{(d) MW-ISPNet \cite{ignatov2020aim}}}
\centerline{\includegraphics[width=\textwidth]{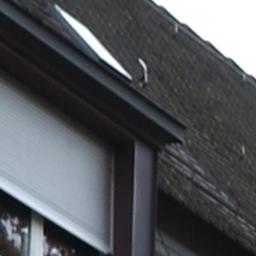}}
\centerline{\scriptsize\textbf{(h) Ground Truth}}
\end{minipage}%
\caption{\label{isp}Visualization comparison of ISP on ZRR \cite{ignatov2020replacing}.}
\vspace{-6pt}
\end{figure}

\textbf{ISP:} On the ZRR \cite{li2019underwater} dataset, the quantitative results in Table \ref{tab:3} show that MobileIE maintains competitive performance while delivering \textbf{3x} faster inference than existing lightweight SOTA models. Figure \ref{isp} provides the corresponding visual comparisons.

\begin{table}[h]
\centering
\vspace{-2pt}
 \resizebox{\columnwidth}{!}{
    \begin{tabular}{cccccc|ccc}
    \toprule
    \multicolumn{1}{c}{Method} & \multicolumn{1}{c}{Venue} & \multicolumn{1}{c}{\#Params$\downarrow$} & \multicolumn{1}{c}{Latency$\downarrow$} & \multicolumn{1}{c}{Latency$\downarrow$} & \multicolumn{1}{c|}{FPS$\uparrow$} & \multicolumn{3}{c}{ZRR \cite{ignatov2020replacing}} \\
    \cmidrule{7-9}
    & & (K)& (GPU,ms) & (SoC,ms) & (448$\times$448) & PSNR$\uparrow$ & SSIM$\uparrow$ & SCORE$\uparrow$\\
    \hline 
    PyNet \cite{ignatov2020replacing} & CVPRW'20 & 47,548& 118.339& \textgreater 500 & \textless 10 & 21.19 & 0.747 & 0.024 \\
    AWNet(raw) \cite{dai2020awnet} & ECCVW'20 & 45,408& 102.932 & \textgreater 500 & \textless 10 & 21.42 & \textcolor{blue}{0.748} & 0.038\\
    MW-ISP \cite{ignatov2020aim} & ECCVW'20 & 29,200& / & / & / & 21.16 & 0.732 & / \\
    LiteISP \cite{zhang2021learning} & ICCV'21 & 11,900 & 46.326 & \textgreater 500 & 21.334 & 21.28 & 0.739 & 0.070\\
    NAFNet \cite{chen2022simple} & ECCV'22 & 7.844 & 3.108 & 78.62 & 321.760 & 21.12 & 0.736 & 0.836 \\
    SYEISP \cite{gou2023syenet} & ICCV'23 & \textcolor{blue}{5.616} & \textcolor{blue}{1.156} & \textcolor{blue}{16.47} & \textcolor{blue}{865.247} & 20.84 & 0.728 & \textcolor{blue}{1.524}\\
    FourierISP \cite{he2024enhancing} & AAAI'24 & 7,590& 22.869 & \textgreater 500 & 43.731 & \textcolor{red}{21.65} & \textcolor{red}{0.755} & 0.237 \\
    \hline
    Ours  & / & \textcolor{red}{4.104} & \textcolor{red}{1.020} & \textcolor{red}{14.40} & \textcolor{red}{980.516} & \textcolor{blue}{21.43} & 0.731 & \textcolor{red}{3.913}\\
    \bottomrule 
    \end{tabular}}
\caption{Performance comparison of different image signal processing models on ZRR datasets. SCORE \cite{ignatov2022learned} represents a comprehensive measure of model performance and efficiency. The top results are marked: best in red and second in blue.}
\label{tab:3}
\vspace{-12pt}
\end{table}

\textbf{Mobile Deployment:} As shown in Tables \ref{tab:1}, \ref{tab:2}, and \ref{tab:3}, MobileIE’s streamlined design (4K parameters and 0.924 GFLOPs) enables seamless deployment on commercial mobile devices, achieving real-time image enhancement at over 100 FPS. This efficiency paves the way for real-time UHD images (2K-8K) enhancement on mobile platforms.

\subsection{Ablation Studies and Analyses}
We conducted a series of ablation studies to verify the effectiveness of MobileIE’s modules. First, MBRConv was compared with other re-parameterization methods \cite{ding2021repvgg,zhang2023repnas,zhang2021edge}, and LVW loss was replaced by conventional loss functions. As shown in Table \ref{tab:4}, both MBRConv and LVW loss significantly improved the model's performance. Importantly, MBRConv merges into standard convolutions during inference, adding no computational or memory overhead.
\begin{table}[htbp]
  \centering
  \resizebox{1\columnwidth}{!}{
    \begin{tabular}{cccccc}
    \toprule
    Datasets & \multicolumn{5}{c}{UIEB \cite{li2019underwater}} \\
    \hline
    Metrics & \multicolumn{1}{c}{PSNR$\uparrow$} & \multicolumn{1}{c}{SSIM$\uparrow$} & \multicolumn{1}{c}{LPIPS$\downarrow$} & \multicolumn{1}{c}{LOE$\downarrow$} & \multicolumn{1}{c}{MAE$\downarrow$} \\
    \hline
    Only inference network & 21.48 & 0.887 & 0.192 & 0.108 & 0.086 \\
    \hline
    L1 loss & 22.20 & 0.902 & 0.168 & 0.098 & 0.078 \\
    L2 loss & 21.74 & 0.894 & 0.175 & 0.095 & 0.083 \\
    Smooth L1 loss & 22.27 & 0.904 & 0.164 & 0.097 & 0.078 \\
    Charbonnier loss & 22.31 & 0.905 & 0.162 & 0.097 & 0.077 \\
    Robust loss \cite{barron2019general} & 22.34 & 0.905 & 0.167 & 0.099 & 0.077 \\
    \hline
    LVW loss & \textbf{22.57} & \textbf{0.906} & \textbf{0.160} & \textbf{0.097} & \textbf{0.075} \\
    \bottomrule
    \bottomrule
    Datasets & \multicolumn{5}{c}{LOLv1 \cite{wei2018deep}+LOLv2-Real \cite{yang2020fidelity}} \\
    \hline
    RepVGG \cite{ding2021repvgg} & 22.69 & 0.821 & 0.202 & 0.257 & 0.089 \\
    RepNAS \cite{zhang2023repnas} & 21.90 & 0.810 & 0.210 & 0.259 & 0.099 \\
    ECBSR \cite{zhang2021edge} & 23.96 & 0.816 & 0.204 & 0.240 & 0.080 \\
    MBRConv(No BN) & 23.27 & 0.821 & 0.232 & 0.258 & 0.082  \\
    MBRConv(No IWO) & 24.02 & 0.823 & 0.199 & 0.257 & 0.078 \\
    \hline
    Ours  & \textbf{24.35} & \textbf{0.829} & \textbf{0.189} & \textbf{0.254} & \textbf{0.074} \\
    \bottomrule
    \end{tabular}}
  \caption{Performance comparison of Loss and Re-param modules.}
  \label{tab:4}
\vspace{-12pt}
\end{table}

\textbf{(1) Why does MBRConv + IWO work?} 

\begin{figure}[htbp] 
\vspace{-4pt}
\begin{minipage}[htbp]{0.74\linewidth}
\centerline{\includegraphics[width=\linewidth]{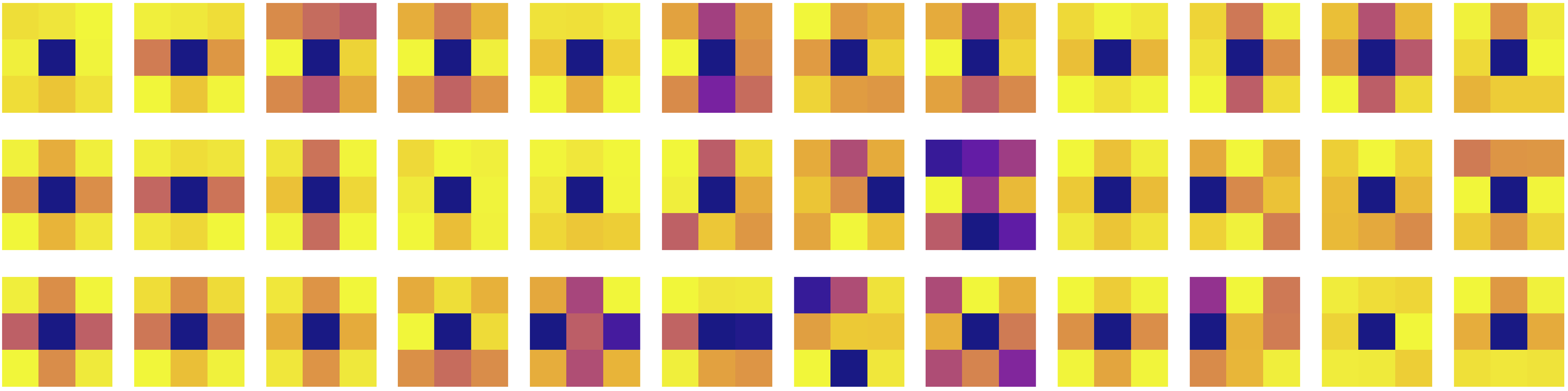}}
\end{minipage}%
\hspace{0.001\linewidth} 
\begin{minipage}[htbp]{0.243\linewidth}
\centerline{\includegraphics[width=\linewidth]{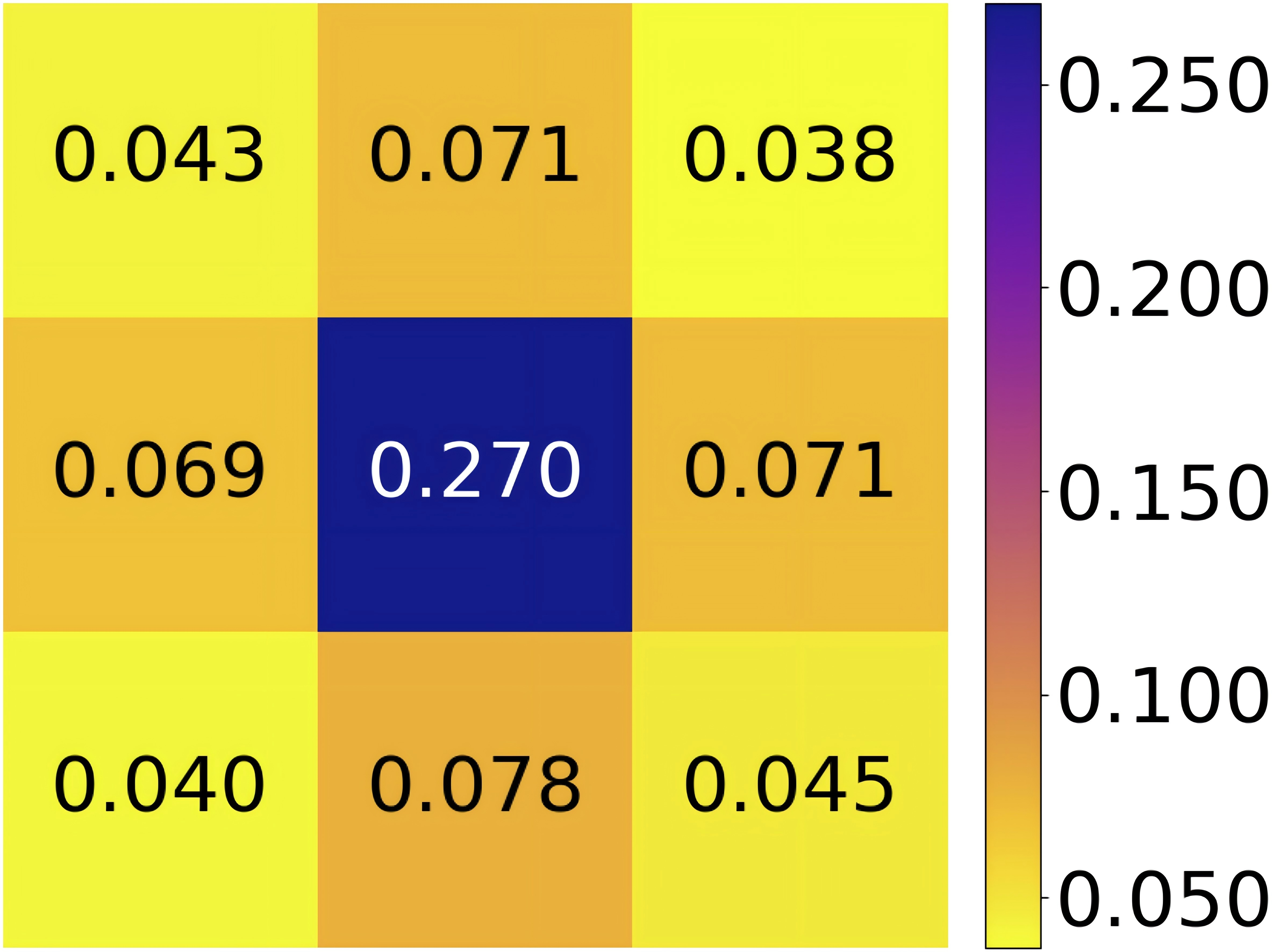}}
\end{minipage}%
\caption{\label{weight} Visualization of kernel differences pre- and post-IWO.}
\vspace{-8pt}
\end{figure}

\textbf{IWO enhances the kernel skeletons.} To explore the impact of IWO on MBRConv kernels, we visualized the weight increments $\Delta W$ in the last convolutional layer of the MobileIE model trained on the UIEB dataset. Here, $\Delta W$ is defined as the difference between the weights trained with IWO ($W_{IWO}$) and those trained without it ($W_{base}$), i.e., $\Delta W = W_{IWO} - W_{base}$. As shown in Figure \ref{weight}, most $\Delta W$ values reinforce the center-row and center-column characteristics of the convolutional kernels, which contributes to improved model performance \cite{ding2019acnet,cai2023refconv}.

\begin{figure}[b]
\vspace{-8pt}
\begin{minipage}[htbp]{0.235\linewidth}
\centerline{\includegraphics[width=\linewidth]{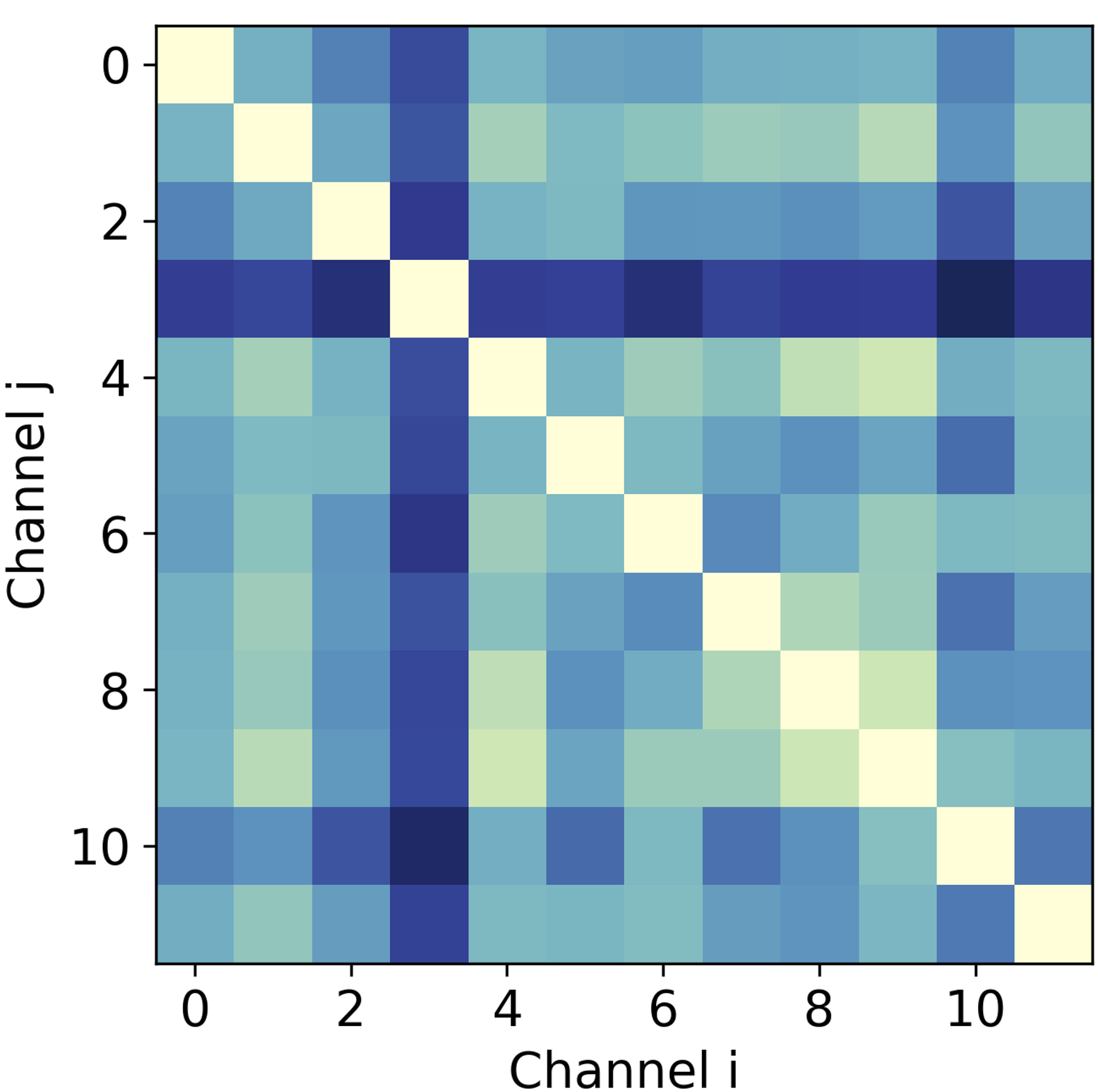}}
\centerline{\scriptsize\textbf{(a) Epoch = 100}}
\end{minipage}%
\begin{minipage}[htbp]{0.235\linewidth}
\centerline{\includegraphics[width=\linewidth]{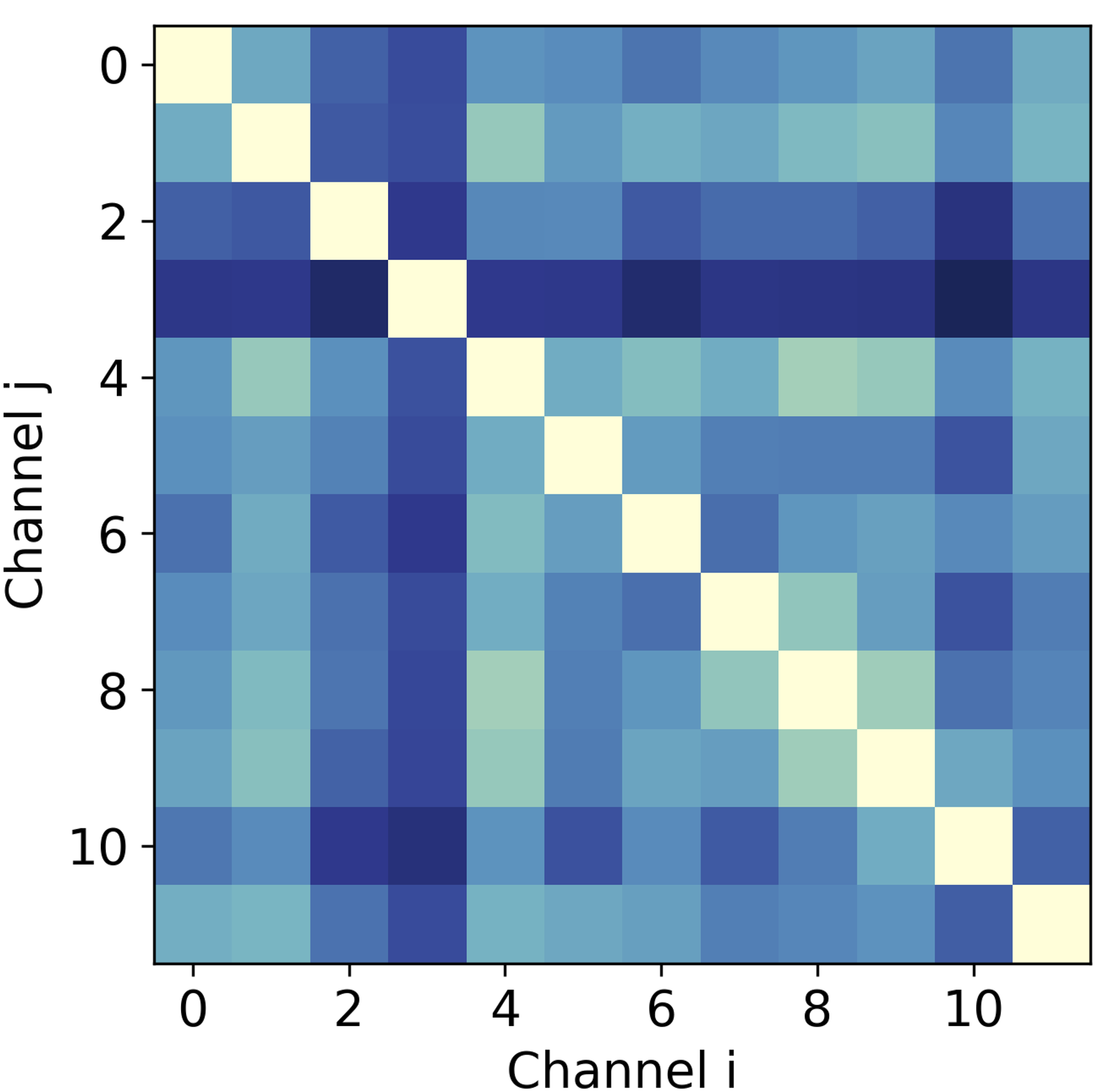}}
\centerline{\scriptsize\textbf{(b) Epoch = 500}}
\end{minipage}%
\begin{minipage}[htbp]{0.235\linewidth}
\centerline{\includegraphics[width=\linewidth]{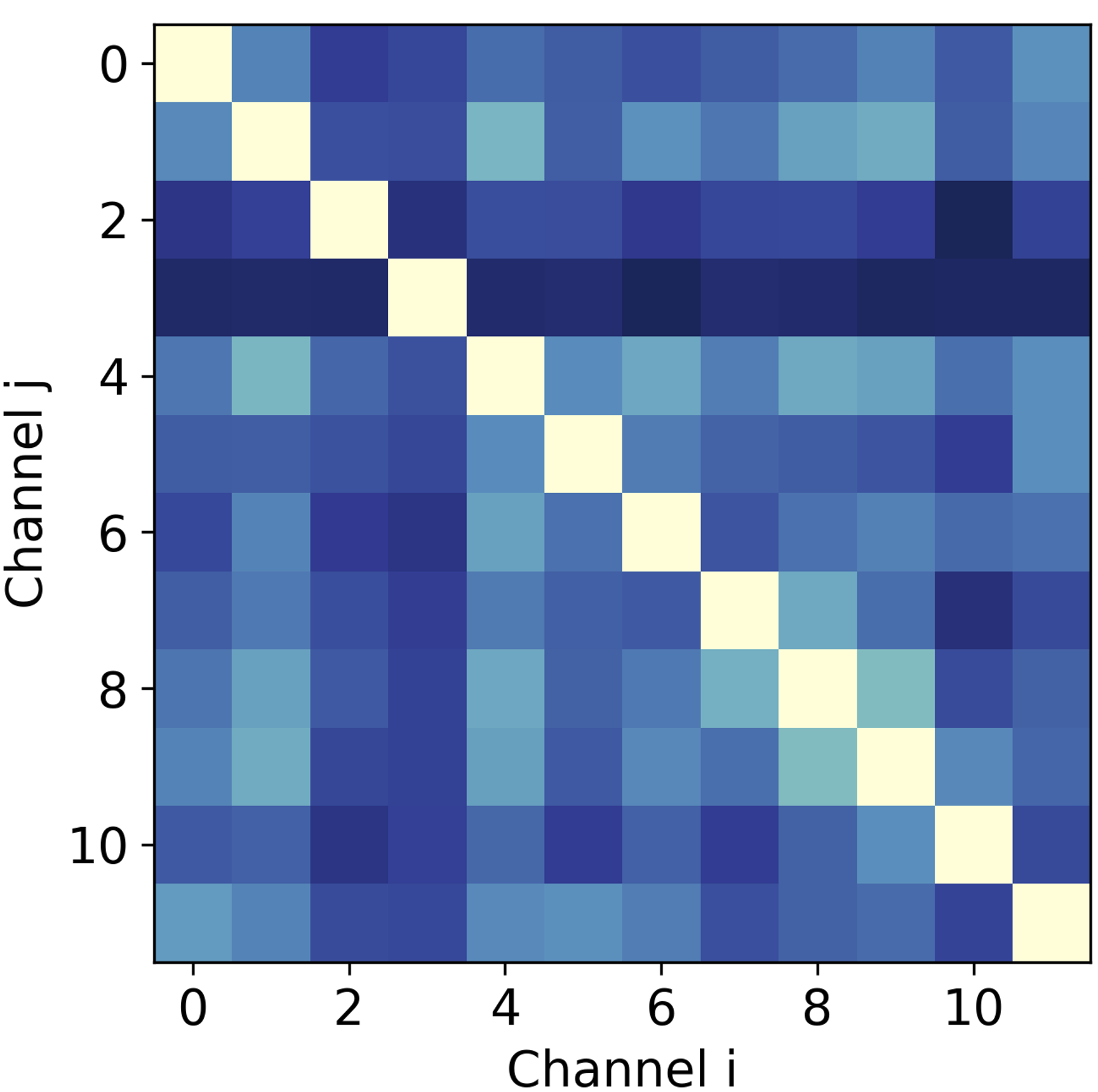}}
\centerline{\scriptsize\textbf{(c) Epoch = 1,000}}
\end{minipage}%
\begin{minipage}[htbp]{0.28\linewidth}
\centerline{\includegraphics[width=\linewidth]{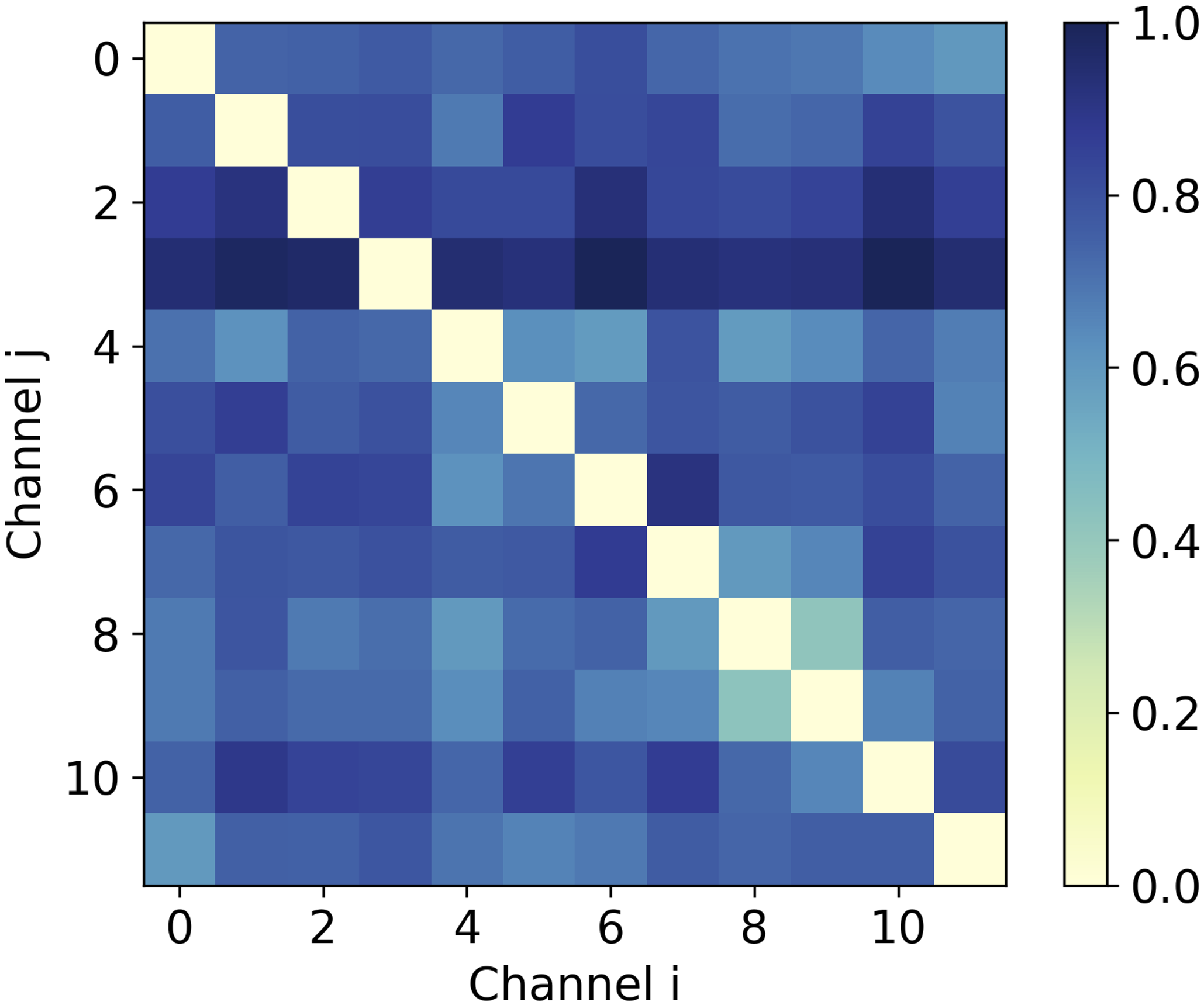}}
\centerline{\scriptsize\textbf{(d) Applying IWO}}
\end{minipage}%
\caption{\label{feature} KL similarity matrix for the last convolutional layer of MBRConv at different training stages.}
\vspace{-8pt}
\end{figure}

\textbf{IWO reduces channel redundancy}, validated by comparing the last Conv$1\times1$ layer of MBRConv$5\times5$ before and after IWO optimization. Channel similarity was measured using Kullback-Leibler (KL) \cite{zhou2019accelerate,wang2021tied,cai2023refconv} divergence, with higher values indicating lower redundancy. We trained MobileIE on the LOLv1, saving checkpoints at different stages. Applying softmax across channels and calculating KL divergence between channel pairs, we generated KL matrices at different training stages (Figure \ref{feature}). As training progressed, KL divergence increased, and after IWO, it rose significantly, underscoring MBRConv’s effectiveness in reducing redundancy and improving feature representation.

\begin{figure}[!h]
\vspace{-14pt}
\centering
\includegraphics[width=\linewidth]{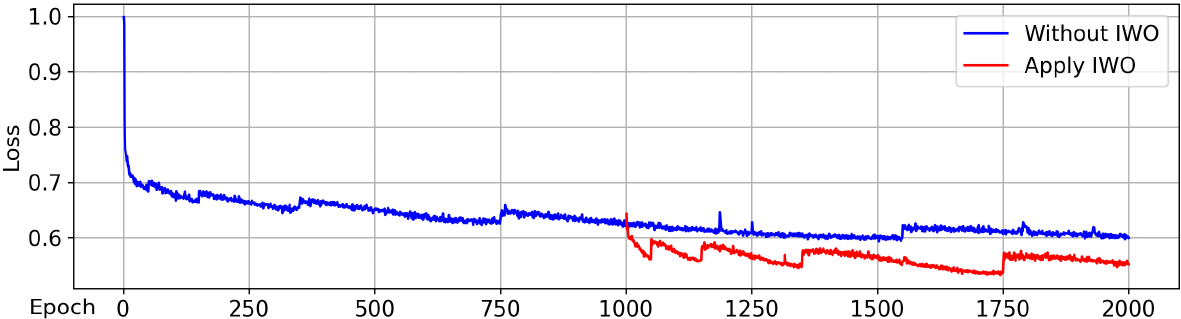}
\caption{\label{lossv} Training loss curves showing the effect of IWO.}
\vspace{-8pt}
\end{figure}

\textbf{Impact of IWO on Training Convergence}. Similar to the Pre-training stage, $W_{pre}$ is the well-performing weight obtained after the first 1000 epochs.  As shown in the training curves (Figure \ref{lossv}), without IWO, the loss stagnates in the later training stages, whereas with IWO, it continues to decrease. This demonstrates that IWO facilitates further optimization of the weights, enhancing the model's convergence performance. The oscillation is due to cosine decay.

\begin{figure}[htbp] 
\begin{minipage}[htbp]{0.238\linewidth}
\centerline{\includegraphics[width=\linewidth]{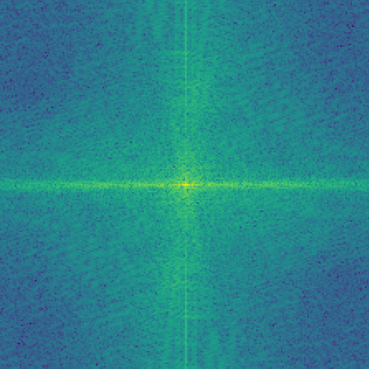}}
\centerline{\scriptsize\textbf{(a) Original Feature:$F$}}
\end{minipage}%
\hspace{0.001\linewidth} 
\begin{minipage}[htbp]{0.238\linewidth}
\centerline{\includegraphics[width=\linewidth]{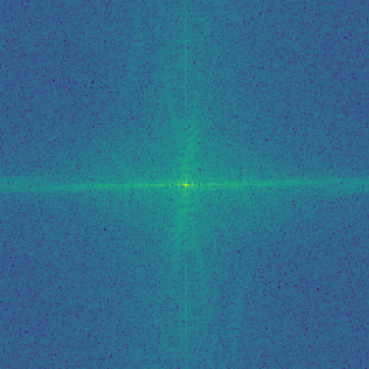}}
\centerline{\scriptsize\textbf{(b) ReLU($F$)}}
\end{minipage}%
\hspace{0.001\linewidth} 
\begin{minipage}[htbp]{0.238\linewidth}
\centerline{\includegraphics[width=\linewidth]{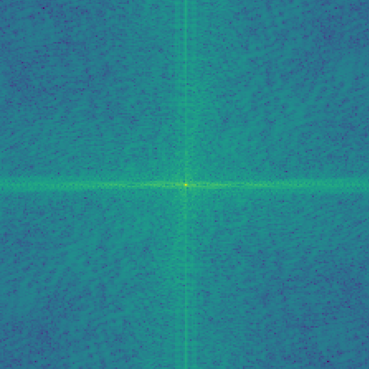}}
\centerline{\scriptsize\textbf{(c) $F\ast F$}}
\end{minipage}%
\hspace{0.001\linewidth} 
\begin{minipage}[htbp]{0.238\linewidth}
\centerline{\includegraphics[width=\linewidth]{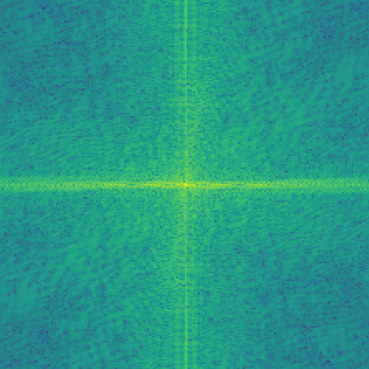}}
\centerline{\scriptsize\textbf{(d) FST($F$)}}
\end{minipage}%
\caption{\label{fusion} Feature Transformation Fourier Spectrum.}
\vspace{-8pt}
\end{figure}

\textbf{(2) FST can enhance nonlinear feature interaction.} 
The FST uses a nonlinear squaring transformation to enhance high-frequency sensitivity, highlighting image details and edges (Figure \ref{fusion}). In the Fourier spectrum, squaring shows a stronger high-frequency response than ReLU, effectively preserving detail. The learnable parameters, $Scale$, and $bias$, further adaptively adjust the feature dynamic range, enhancing restoration performance.

\begin{figure}[h]
\centering
\includegraphics[width=0.97\linewidth]{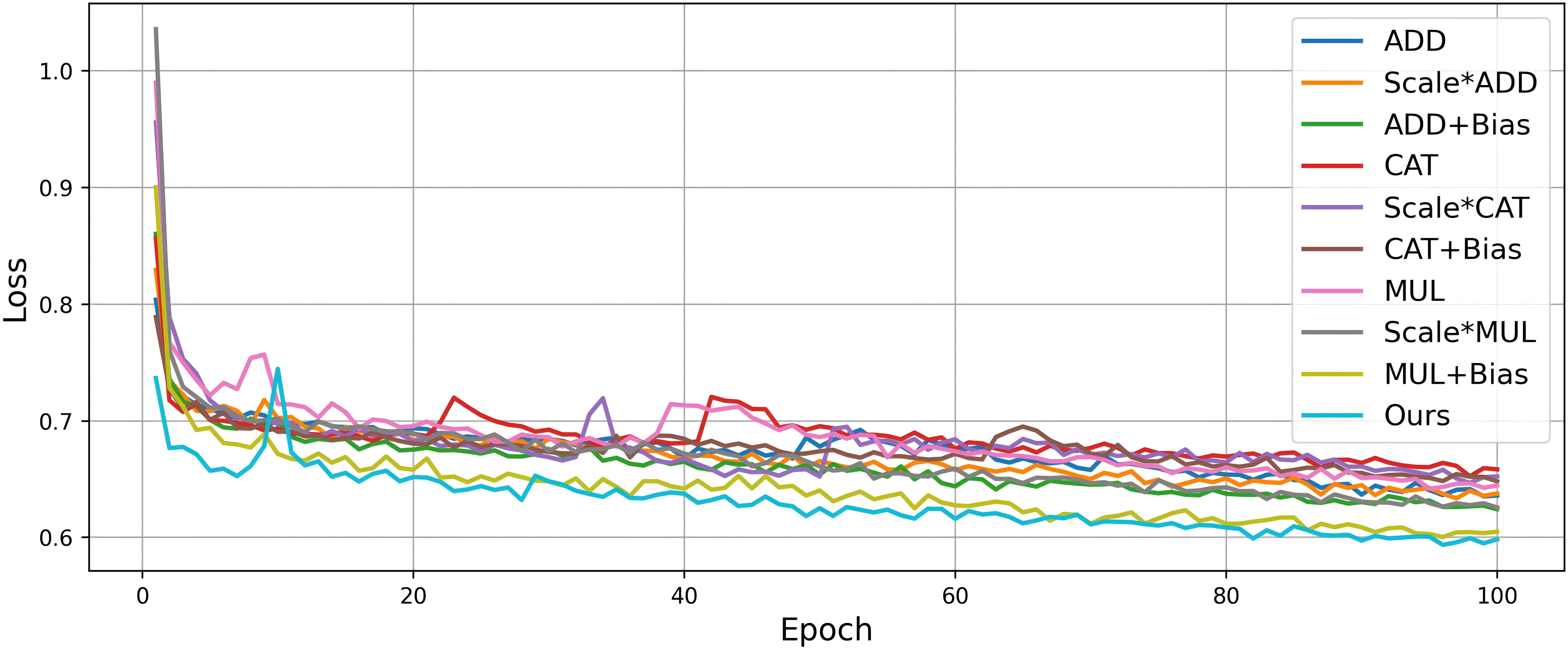}
\caption{Training loss comparison of different feature transform.}
\label{loss1}
\vspace{-8pt}
\end{figure}

Figure \ref{loss1} shows that the FST achieves a faster convergence rate during training compared to other feature transform methods. In the initial epochs, the FST rapidly reduces loss, demonstrating its ability to enhance feature representation early and accelerate convergence. Additionally, the FST’s final stabilized loss is lower than other methods, highlighting its robustness and effectiveness in complex tasks. Table \ref{tab:5} further confirms that FST delivers the best feature transform performance on the UIEB \cite{li2019underwater} dataset.

\begin{table}[htbp]
  \centering
  \resizebox{0.85\columnwidth}{!}{
    \begin{tabular}{cccccc}
    \toprule
    Methods & PSNR$\uparrow$ & SSIM$\uparrow$ & LPIPS$\downarrow$ & LOE$\downarrow$ & MAE$\downarrow$ \\
    \midrule
    Baseline & 21.12  & 0.884  & 0.199  & 0.101  & 0.088  \\
    ADD   & 21.50  & 0.896  & 0.178  & 0.107  & 0.084  \\
    Scale$\ast$ADD & 21.34  & 0.893  & 0.173  & 0.105  & 0.087  \\
    ADD+Bias & 21.58  & 0.890  & 0.177  & 0.093  & 0.086  \\
    CAT   & 21.44  & 0.893  & 0.181  & 0.095  & 0.085  \\
    MUL   & 21.80  & 0.901  & 0.173  & 0.095  & 0.084  \\
    MUL+Bias & 21.84  & 0.901  & 0.168  & 0.092  & 0.081  \\
    Scale$\ast$MUL & 21.95  & 0.902  & 0.167  & 0.092  & 0.081  \\
    \hline
    Ours & \textbf{22.60} & \textbf{0.906} & \textbf{0.157} & \textbf{0.092} & \textbf{0.075}  \\
    \bottomrule
    \end{tabular}}
  \caption{Performance comparison of different feature transform. "Baseline" represents the model without any feature transform.}
  \label{tab:5}
\vspace{-8pt}
\end{table}

\textbf{(3) Dual-path fusion enables precise feature capture.} 

How can one achieve both efficiency and precise capture of local and global feature dependencies? Following Occam’s Razor principle, HDPA adopts a streamlined design language, utilizing a dual-path structure and adaptive feature interaction to decompose and aggregate features hierarchically, thereby extracting critical information across different scales. This dual-path design also supports mutual optimization during backpropagation.

\begin{figure}[htbp] 
\begin{minipage}[htbp]{0.243\linewidth}
\centerline{\includegraphics[width=\linewidth]{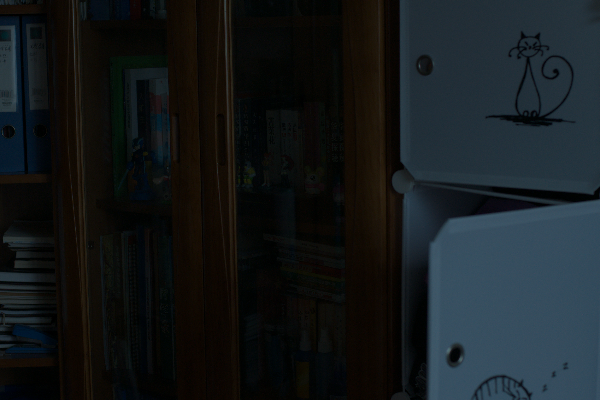}}
\centerline{\scriptsize\textbf{(a) RGB image}}
\end{minipage}%
\hspace{0.00001\linewidth} 
\begin{minipage}[htbp]{0.243\linewidth}
\centerline{\includegraphics[width=\linewidth]{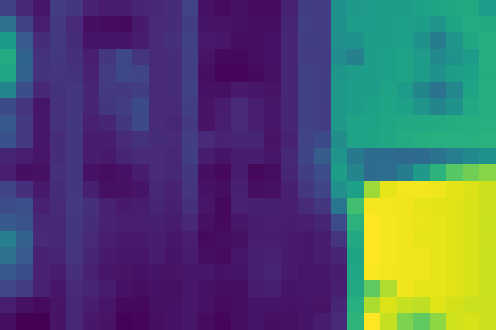}}
\centerline{\scriptsize\textbf{(b) Input feature}}
\end{minipage}%
\hspace{0.00001\linewidth} 
\begin{minipage}[htbp]{0.243\linewidth}
\centerline{\includegraphics[width=\linewidth]{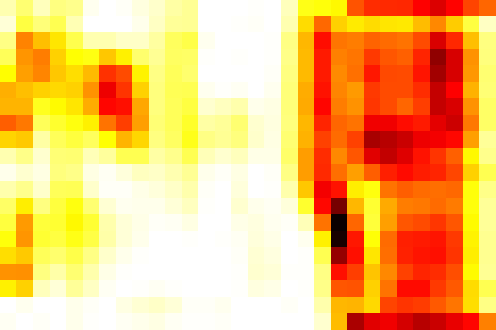}}
\centerline{\scriptsize\textbf{(c) HDPA heat map}}
\end{minipage}%
\hspace{0.00001\linewidth} 
\begin{minipage}[htbp]{0.243\linewidth}
\centerline{\includegraphics[width=\linewidth]{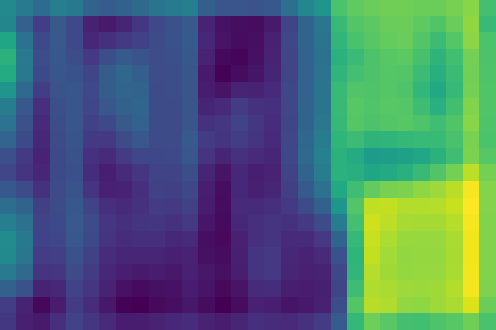}}
\centerline{\scriptsize\textbf{(d) Output feature}}
\end{minipage}%
\caption{\label{att} Visualization of feature maps at different stages.}
\vspace{-8pt}
\end{figure}

Figure \ref{att} illustrates HDPA’s effectiveness. The initial convolution preserves key edges and structures, while the HDPA heatmap highlights important regions, focusing attention on key details. This produces a sharper output feature map, demonstrating how HDPA’s attention improves image enhancement by refining key details.

\begin{table}[htbp]
\centering
 \resizebox{\columnwidth}{!}{
    \begin{tabular}{ccccc|cccc}
    \toprule
    \multicolumn{1}{c}{Method} & \multicolumn{1}{c}{\#Params$\downarrow$} & \multicolumn{1}{c}{FLOPs$\downarrow$} & \multicolumn{1}{c}{Latency$\downarrow$} & \multicolumn{1}{c|}{Latency$\downarrow$} & \multicolumn{4}{c}{LOLv1 \cite{ignatov2020replacing}} \\
    \cmidrule{6-9}
    & (K) & (G) & (GPU,ms) & (SoC,ms) & PSNR$\uparrow$ & SSIM$\uparrow$ & LPIPS$\downarrow$ & SCORE$\uparrow$ \\
    \midrule 
    SE-Net\cite{hu2018squeeze} & 4.179 & 0.922 & 0.948 & 7.41 & 22.27 & 0.804 & 0.217 & 17.259\\
    CBAM\cite{woo2018cbam}  & 4.254 & 0.945 & 6.573 & 95.71 & 22.38 & 0.796 & 0.231 & 1.556\\
    ECA-Net\cite{wang2020eca} & \textbf{3.870} & 0.922 & 0.805  & 6.65 & 21.79 & 0.799 & 0.221 & 9.886 \\
    EA\cite{shen2021efficient} & 4.491 & 1.057 & 1.232 &  11.45 & 19.93 & 0.779 & 0.260 & 0.436\\
    NAM\cite{liu2021nam}   & 3.891 & 0.930 & 0.905 & 7.02 & 20.10 & 0.761 & 0.252 & 0.900\\
    SCA\cite{chen2022simple} & 4.023 & 0.922 & \textbf{0.769} & 6.33 & 20.82 & 0.786 & 0.250 & 2.707\\
    DFC\cite{tang2022ghostnetv2}   & 3.979 & \textbf{0.921} & 0.890  & \textbf{6.29} & 21.50 & 0.802 & 0.208 & 6.992\\
    EMA\cite{ouyang2023efficient}   & 4.239 & 1.086 & 1.569 & 9.79 & 20.78 & 0.792 & 0.233 & 1.656\\
    SPAN\cite{wan2024swift}  & 5.511 & 1.308 & 1.835 & 10.24 & 19.79 & 0.767 & 0.260 & 0.401\\
    CGA\cite{chen2024dea}   & 6.330 & 1.486 & 1.437 & 9.62 & 22.33 & 0.803 & 0.209 & 14.447\\
    LKA\cite{azad2024beyond}   & 5.247 & 1,236 & 1.082 & 8.96 & 19.76 & 0.782 & 0.227 & 0.440\\
    \midrule 
    Ours & 4.035 & 0.924 & 0.886 & 6.72 & \textbf{23.26} & \textbf{0.805} & \textbf{0.213} & \textbf{75.079}\\
    \bottomrule
    \end{tabular}}
\caption{Performance comparison of different Attention mechanisms on LOLv1 \cite{wei2018deep} dataset. SCORE \cite{ignatov2022learned} reflects the efficiency.}
  \label{tab:6}
\vspace{-8pt}
\end{table}

Finally, we replaced it with mainstream attention mechanisms and conducted ablation experiments, as shown in Table \ref{tab:6}.  With minimal added parameters (+0.168 K) and FLOPs (+5.76 M), HDPA achieved a PSNR gain of +3.08 dB. Tests on mobile devices confirmed HDPA’s high performance with minimal latency, underscoring the efficiency of our edge-focused IE design in MobileIE.

\begin{figure}[htbp] 
\begin{minipage}[htbp]{0.5\linewidth}
\centerline{\includegraphics[width=\linewidth]{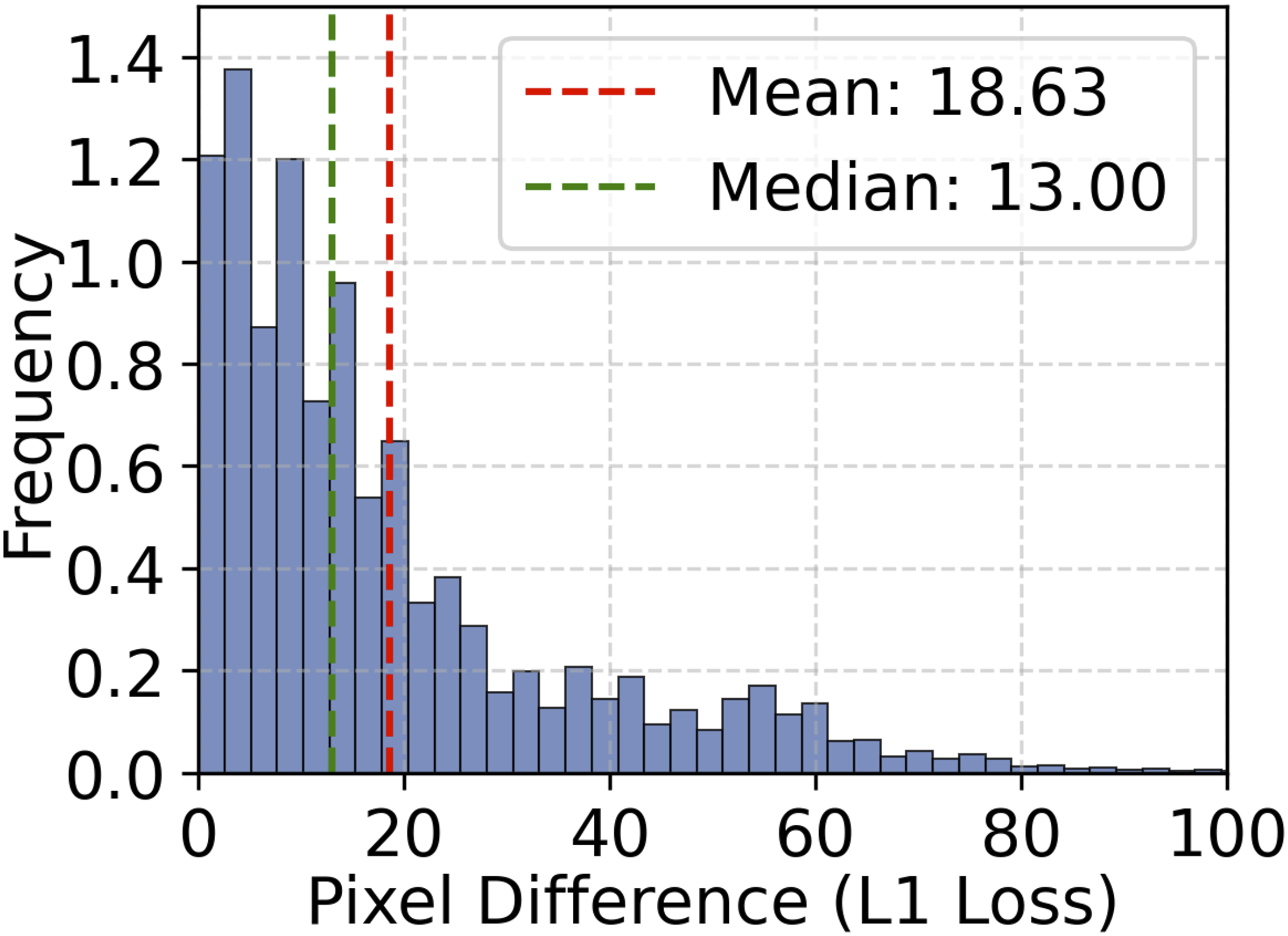}}
\centerline{\includegraphics[width=\linewidth]{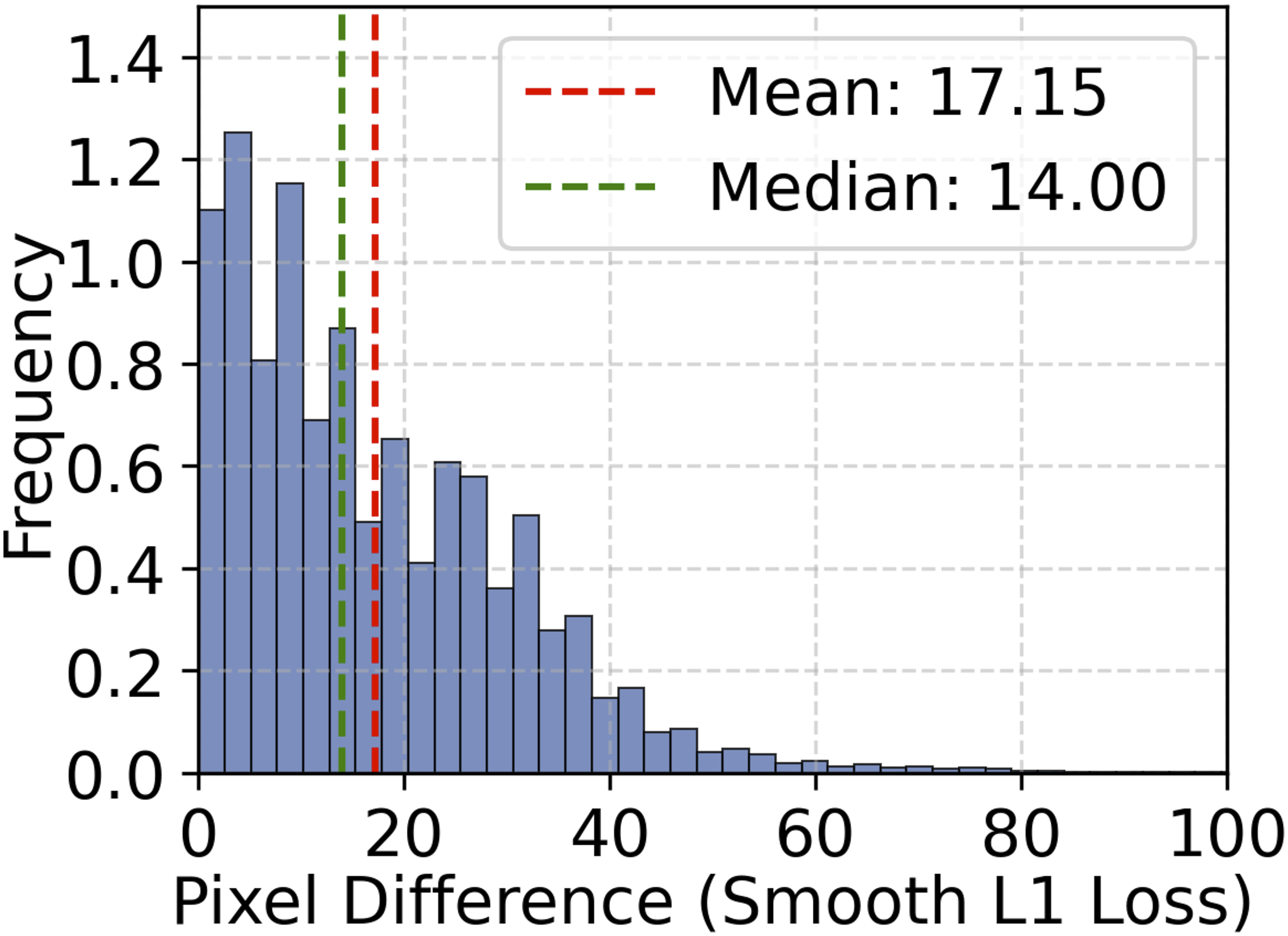}}
\end{minipage}%
\begin{minipage}[htbp]{0.5\linewidth}
\centerline{\includegraphics[width=\linewidth]{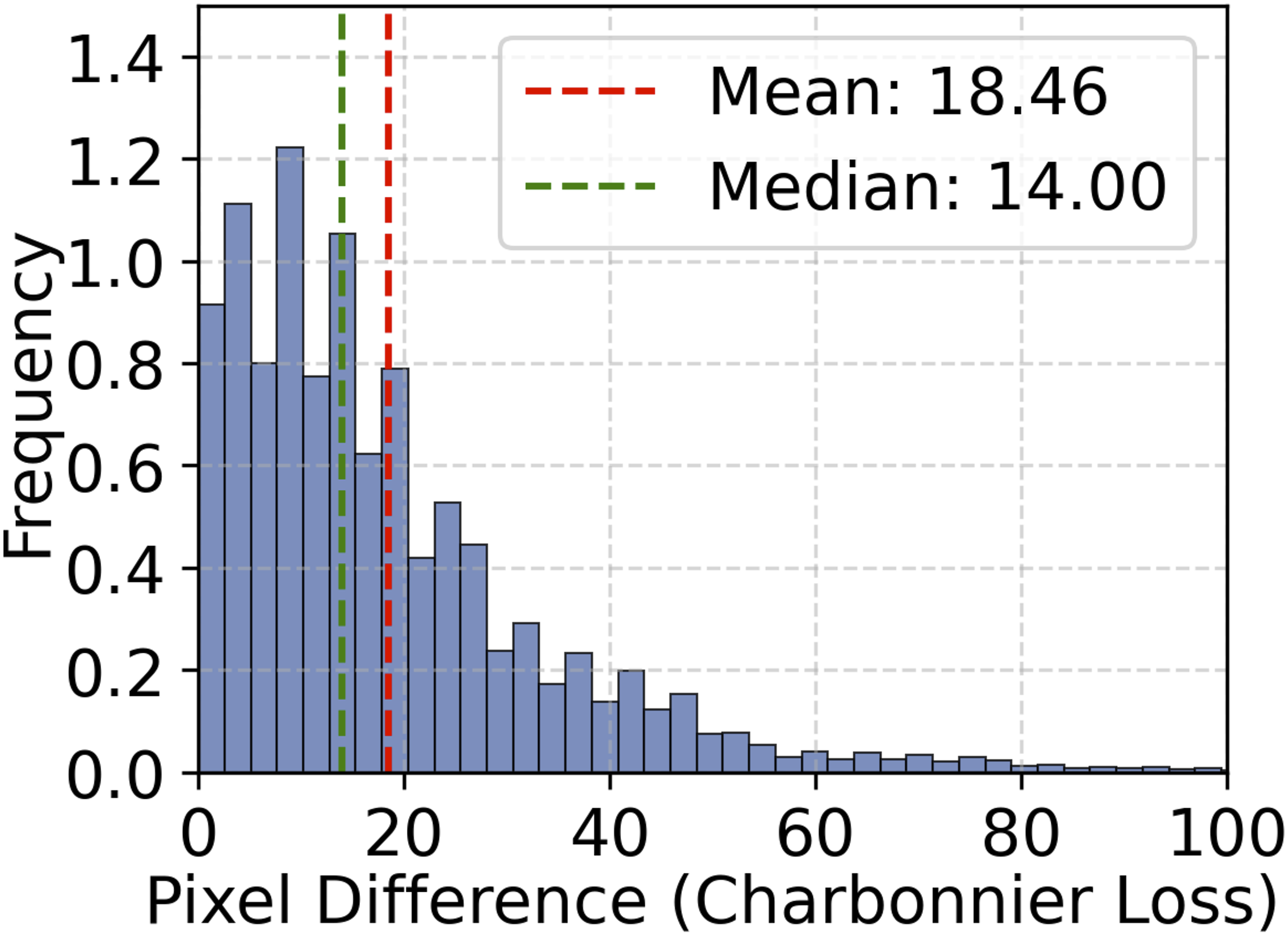}}
\centerline{\includegraphics[width=\linewidth]{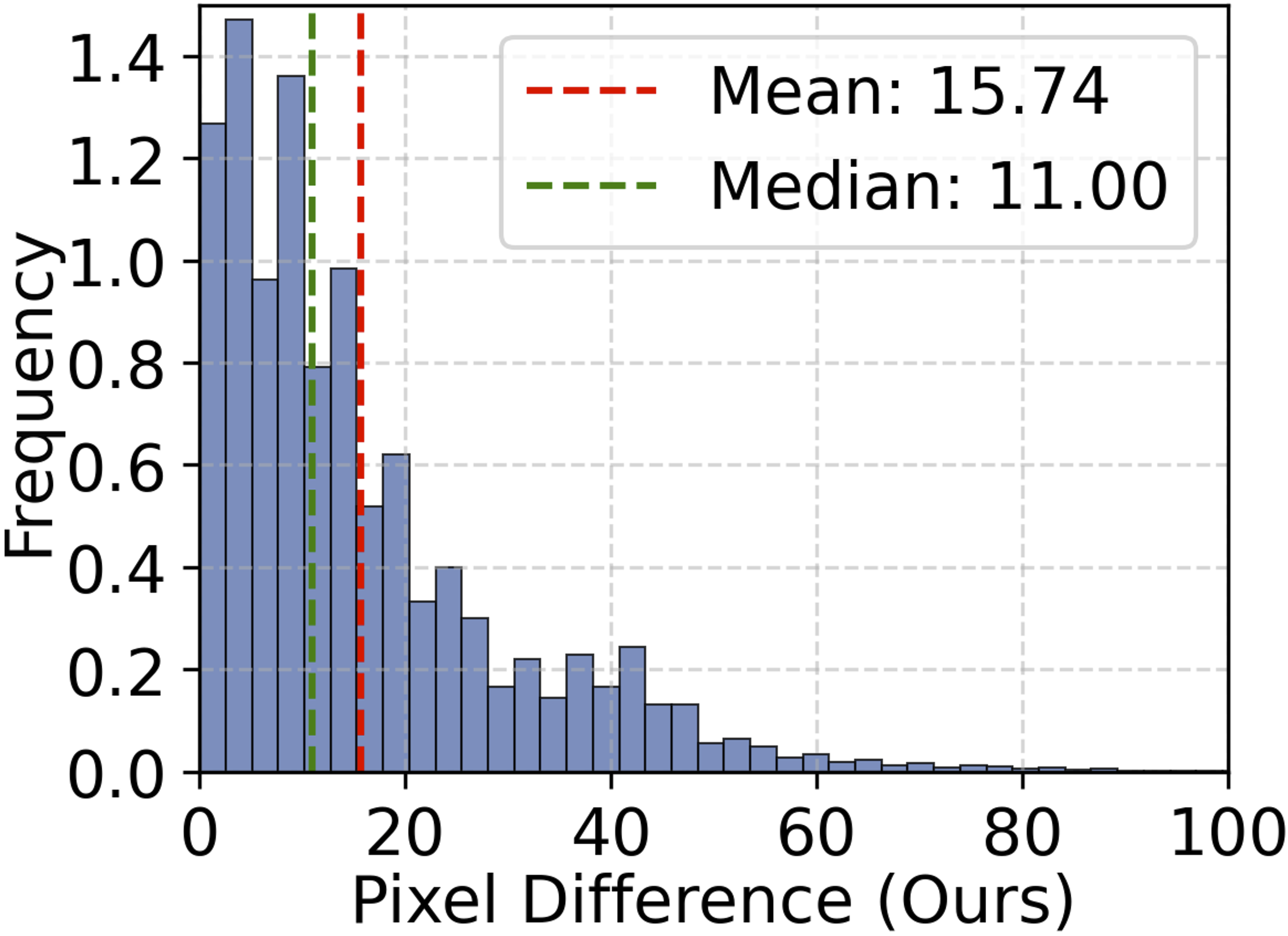}}
\end{minipage}%
\caption{\label{losshis} Visualization of pixel prediction error histograms for various loss functions.}
\vspace{-8pt}
\end{figure}

\textbf{(4) Effectiveness of LVW in Outlier Optimization.} 
We observed that the pixel error histograms follow a Laplace-like distribution, suggesting the model optimizes accurate pixels while neglecting outliers \cite{gou2023syenet}. Based on L1 Loss, LVW adjusts weights to increase the constraint on outliers. Normalization standardizes errors, bringing the data distribution closer to uniform and ensuring balanced pixel weights. Additionally, we visualized the pixel prediction error histograms for different loss functions, as shown in Figure \ref{losshis}. LVW exhibits lower mean and median pixel errors, improving effectiveness in optimizing challenging regions.

\textbf{(5) Module-Wise Lightweight Contributions.} Table \ref{tab:7} demonstrates that re-parameterization significantly reduces model complexity during inference. IWO and LVW introduce no extra cost, making them suitable for efficient deployment. Re-parameterization combined with cost-free strategies enables efficient and lightweight inference.

\begin{table}[!h]
  \centering
  \vspace{-6pt}
  \resizebox{\columnwidth}{!}{
    \begin{tabular}{cccc|ccc|cc}
    \toprule 
    Model & Training & w/o FST & w/o HDPA & Inference & w/o FST & w/o HDPA & IWO & LVW \\
    \toprule
    Params (K)      & 75.243 & 75.243 & 71.283 & 4.047 & 4.047 & 3.867 & & \\
    Model Size (MB) & 0.29   & 0.29   & 0.27   & 0.02  & 0.02  & 0.01  & \multicolumn{2}{c}{Cost-Free} \\
    FLOPs (G)       & 17.447 & 17.447 & 17.099 & 0.924 & 0.924 & 0.919 & \multicolumn{2}{c}{Optimization} \\
    Latency (ms)    & 12.83  & 12.61  & 11.31  & 0.895 & 0.864 & 0.732 & & \\
    \bottomrule 
    \end{tabular}
  }
  \caption{Ablation on Module Efficiency (RTX 3090).}
  \label{tab:7}
  \vspace{-13pt}
\end{table}
\section{Conclusion}
In this paper, we introduce MobileIE, a lightweight CNN with only 4K parameters designed for real-time image enhancement on mobile devices. MobileIE combines simplicity with high performance across low-light image enhancement, underwater image enhancement, and image signal processing tasks. Key components such as the MBRConv, Feature Self-Transform module, Hierarchical Dual-Path Attention mechanism, and Local Variance-Weighted loss enable MobileIE to achieve superior results while maintaining an inference time of around 0.9 ms. Furthermore, MobileIE can be easily deployed on any commercial mobile device with minimal computational overhead and efficient resource utilization. Future work will explore optimizations to expand MobileIE’s application to more different tasks.

{
    \small
    \bibliographystyle{ieeenat_fullname}
    \bibliography{main}
}

\end{document}